\documentclass[sigconf]{acmart}
\AtBeginDocument{%
  }

\copyrightyear{2025} 
\acmYear{2025} 
\setcopyright{cc}
\setcctype{by}
\acmConference[KDD '25]{Proceedings of the 31st ACM SIGKDD Conference on
Knowledge Discovery and Data Mining V.2}{August 3--7, 2025}{Toronto, ON,
Canada}
\acmBooktitle{Proceedings of the 31st ACM SIGKDD Conference on Knowledge
Discovery and Data Mining V.2 (KDD '25), August 3--7, 2025, Toronto, ON,
Canada}
\acmDOI{10.1145/3711896.3737096}
\acmISBN{979-8-4007-1454-2/2025/08}

\usepackage{algorithm}
\usepackage{algorithmic}
\usepackage{amsmath}
\usepackage{amsthm}  
\usepackage{subcaption}
\usepackage{booktabs}
\usepackage{multicol}
\usepackage{multirow}
\usepackage{pifont}
\usepackage{xcolor}
\usepackage{tcolorbox}
\usepackage{tikz}
\usepackage{caption}
\usepackage{threeparttable}
\usepackage{colortbl}
\usepackage[symbol]{footmisc}  
\usepackage{balance}

\newcommand{\stitle}[1]{\vspace{1mm}\noindent\textbf{#1}}

\begin{document}

\title{Quantizing Text-attributed Graphs for Semantic-Structural Integration}

\author{Jianyuan Bo}
\affiliation{%
  \institution{Singapore Management University}
  \city{Singapore}
  \country{Singapore}}
\email{jybo.2020@phdcs.smu.edu.sg}

\author{Hao Wu}
\authornote{Corresponding authors.}
\affiliation{%
  \institution{Beijing Normal University}
  \city{Beijing}
  \country{China}}
\email{wuhao@bnu.edu.cn}

\author{Yuan Fang}
\authornotemark[1]  
\affiliation{%
  \institution{Singapore Management University}
  \city{Singapore}
  \country{Singapore}}
\email{yfang@smu.edu.sg}


\renewcommand{\shortauthors}{Jianyuan Bo, Hao Wu, and Yuan Fang}

\begin{abstract}
Text-attributed graphs (TAGs) have emerged as a powerful representation for modeling complex relationships across diverse domains. With the rise of large language models (LLMs), there is growing interest in leveraging their capabilities for graph learning. However, current approaches face significant challenges in embedding structural information into LLM-compatible formats, requiring either computationally expensive alignment mechanisms or manual graph verbalization techniques that often lose critical structural details. Moreover, these methods typically require labeled data from source domains for effective transfer learning, significantly constraining their adaptability. We propose STAG, a novel self-supervised framework that directly quantizes graph structural information into discrete tokens using a frozen codebook. Unlike traditional quantization approaches, our method employs soft assignment and KL divergence guided quantization to address the unique challenges of graph data, which lacks natural tokenization structures. Our framework enables both LLM-based and traditional learning approaches, supporting true zero-shot transfer learning without requiring labeled data even in the source domain. Extensive experiments demonstrate state-of-the-art performance across multiple node classification benchmarks while maintaining compatibility with different LLM architectures, offering an elegant solution to bridging graph learning with LLMs. 
\end{abstract}

\ccsdesc[500]{Computing methodologies~Artificial intelligence}
\ccsdesc[500]{Computing methodologies~Unsupervised learning}
\ccsdesc[300]{Computing methodologies~Neural networks}

\keywords{text-attributed graphs, large language models, few-shot, zero-shot, graph transfer learning}



\maketitle

\section{Introduction}
Graphs serve as a cornerstone for modeling and understanding complex relationships across diverse domains, from social media~\cite{perozzi2014deepwalk,grover2016node2vec,kipf2016semi} and knowledge graphs~\cite{bordes2013translating,wang2014knowledge,trouillon2016complex} to recommendation systems~\cite{fan2019graph,he2020lightgcn,wang2019neural}. The structural information inherent in graphs is critical for effective graph learning, driving the development of graph neural networks (GNNs)~\cite{velivckovic2018graph}. Meanwhile, many real-world graphs contain textual descriptions, such as paper abstracts in citation networks~\cite{yang2016revisiting}, and product descriptions in co-purchase networks~\cite{mcauley2015image}. 
Such graphs are known as \emph{text-attributed graphs} (TAGs)~\cite{wen2024prompt,yan2023comprehensive}, in which nodes or edges are associated with rich textual content. However, conventional GNNs struggle to effectively utilize raw text in TAGs. With the rise of pre-trained language models (PLMs)~\cite{kenton2019bert,he2020deberta} and large language models (LLMs)~\cite{brown2020language,touvron2023llama,achiam2023gpt,hoffmann2022training,yang2024qwen2,guo2025deepseek}, there is growing interest in combining LLMs with graph learning, termed as \emph{GraphLLM}~\cite{chen2024exploring}. Among existing GraphLLM studies, processing TAGs represents an important direction due to the abundance of both semantic and structural information in TAGs. 
Particularly, text attributes can be processed by LLMs to obtain semantically rich initial features, offering significant advantages over traditional shallow features.

\stitle{Limitations of Prior Work.} Despite the potential of GraphLLM in processing TAGs, a fundamental challenge remains: how to effectively unify both semantic and structural information, integrating them into formats that LLMs can utilize. The challenges arise from the misalignment between continuous embedding spaces used for structural encoding and the discrete token spaces native to LLMs. As a result, current approaches particularly struggle with integrating neighborhood structures into the textual semantics in LLM-compatible formats. 
Graph prompting approaches \cite{liu2023one} primarily use LLMs for semantic feature extraction, then manually construct prompt nodes or ego-networks to capture neighborhood structures, which are subsequently encoded using GNNs. However, they fail to leverage the powerful inference capabilities of LLMs.
Hence, some methods have sought a deeper integration of structure-based GNN embeddings with LLMs, relying on computationally expensive and LLM-specific projectors for alignment~\cite{tang2024graphgpt}.
Additionally, such methods often resort to manually designed subgraph verbalization techniques to describe the neighborhood to LLMs (e.g., ``central node A, linked to B and C'')~\cite{ye2023natural,zhao2023graphtext}, which not only hinders scalability to large neighborhoods but also introduces inconsistency and instability across different LLMs~\cite{guo2023gpt4graph}.

The second challenge lies in cross-dataset transfer learning. 
To overcome negative transfer and feature misalignment across different datasets~\cite{sun2023all}, additional labeled data in source or target domains are often used. ZeroG~\cite{li2024zerog} fine-tunes PLMs to align structurally-aware embeddings with semantic features based on labeled data in the source datasets. Similarly, OFA~\cite{liu2023one} requires training with labeled data from various domains to enable effective transfer learning.
The labeling requirement can be costly and significantly constrain adaptability across diverse downstream tasks, while supervised fine-tuning of PLMs is computationally expensive and offers limited generalization across domains~\cite{li2024zerog}.

\stitle{Proposed Work.} These challenges point to a critical need for a fundamentally different approach to embedding structural information of TAGs in GraphLLMs. Directly mapping graph embeddings into discrete token spaces would eliminate the need for expensive alignment mechanisms and manual graph verbalization, enabling native compatibility across different LLMs while integrating both structural and semantic information.

Inspired by vector quantization techniques like VQ-VAE~\cite{van2017neural}, which quantize images into discrete tokens for LLM processing~\cite{zhu2024beyond}, we address the challenges of GraphLLM through a novel approach that fundamentally shifts from continuous embeddings to discrete tokens for TAGs. Our key insight is using quantization techniques to encode nodes into discrete tokens that seamlessly integrate structural and semantic information.
However, quantizing graph structural information presents unique opportunities that distinguish our work from quantization in computer vision. First, unlike in computer vision, where feature maps provide natural structures for tokenization~\cite{zhu2024beyond}, graphs lack inherent patterns for encoding neighborhood information to achieve semantic-structural integration.
Second, na\"ive quantization approaches risk overfitting to pre-training datasets, potentially hindering transferability. Overfitting is less common in computer vision, which benefits from large-scale, diverse pre-training datasets~\cite{deng2009imagenet,lin2014microsoft}. In contrast, graph domains often lack such comprehensive pre-training data, making cross-dataset transfer challenging without additional labeled data.

To overcome these issues in the quantization of TAGs, we  propose a framework called \textbf{S}oft \textbf{T}okenization for Text-\textbf{a}ttributed \textbf{G}raphs (STAG). During pre-training, a GNN first learns node representations that capture structural information. These structure-based node embeddings are then quantized into discrete tokens with self-supervised learning objectives: (1) a reconstruction loss for preserving semantic information, and (2) a contrastive loss for capturing neighborhood structural information. This design ensures effective preservation of both semantic and structural information without requiring labeled data. 
More specifically, we design a \emph{soft assignment strategy} to map each node to a distribution of tokens,
compensating for the lack of explicit tokenization structures in graphs. In contrast, computer vision employs a hard assignment, where each feature map or patch is naturally mapped to a single token. The soft assignment also prevents overfitting to specific tokens and thus improves transferability across domains. Furthermore, we incorporate a Kullback-Leibler (KL) divergence loss to guide the quantization process, enhancing the alignment between structural and semantic representations in the absence of labeled data. 

During inference, STAG can flexibly work with LLMs by providing quantized tokens as prompts in zero- or few-shot settings, or function independently by using the learned embeddings through prompt tuning or linear probing in few-shot settings. It also supports both single- and cross-dataset learning, requiring no source labels in the latter, unlike previous approaches \cite{liu2023one}.

\stitle{Contributions.} Our key contributions advance the state of the art in GraphLLM in three significant ways. (1) We propose a novel quantization approach for TAGs that bridges the gap between continuous graph embeddings and discrete LLM token spaces. (2) We develop a unified framework that supports diverse learning paradigms with or without LLMs in both single- and cross-dataset scenarios. In particular, in cross-dataset learning, we achieve true zero-shot learning without requiring any labeled data for the source dataset. (3) We conduct extensive empirical validation, demonstrating the superior performance of STAG across multiple graph benchmarks. 

\section{Related Work}

In this section, we briefly review related work on GraphLLM and vector quantization. 

\stitle{Graph Learning with LLMs.} Recent advances in large language models (LLMs) have inspired several approaches to combine LLMs with graph learning. GraphGPT~\cite{tang2024graphgpt} converts graph structures into natural language descriptions for LLM processing, while GIMLET~\cite{zhao2023gimlet} proposes a unified graph-text language model for zero-shot learning, though both either lose structural information or require extensive architectural modifications. OFA~\cite{liu2023one} and Pro\-digy~\cite{huang2024prodigy} construct prompts to leverage PLMs' semantic representations, but struggle to fully utilize LLMs' capabilities. While GraphText~\cite{zhao2023graphtext} attempts to bridge graph-LLM semantic spaces through learned projections, it requires computationally expensive alignment procedures. TAPE~\cite{he2023harnessing} takes a different approach by using LLM-generated explanations as node features for GNNs, achieving state-of-the-art performance efficiently. Despite these advances, fundamental challenges persist in preserving both structural and semantic information while maintaining cross-architecture compatibility. Our work addresses these limitations through a novel quantization-based approach that enables direct processing of graph data by frozen LLMs without compromising structural integrity.

\stitle{Graph-to-LLM Representation.} The challenge of representing graphs for LLM processing has led to diverse solutions. Traditional approaches rely on graph verbalization~\cite{ye2023natural,zhao2023graphtext}, which faces scalability issues with large graphs. More structured representations have emerged through code-like formats and adjacency tables~\cite{guo2023gpt4graph,fatemi2023talk,wang2024can}, though they struggle to balance structural completeness with LLM compatibility. Recent embedding-fusion methods~\cite{he2024g,tian2024graph} better preserve structural information by integrating GNN embeddings with LLM representations, but require expensive alignment mechanisms. Position-aware approaches offer another perspective, with GIMLET~\cite{zhao2023gimlet} using generalized position embeddings and LINKGPT~\cite{he2024linkgpt} employing pairwise encoding to capture structural relationships. However, these methods face challenges in cross-architecture consistency and computational efficiency at scale.

\stitle{Vector Quantization.} Vector quantization has evolved significantly since the introduction of VQ-VAE~\cite{van2017neural}, which pioneered neural discrete representation learning. Subsequent works have enhanced this framework through hierarchical structures in VQ-VAE-2~\cite{razavi2019generating}, adversarial training in VQ-GAN~\cite{esser2021taming}, and residual quantization in RQ-VAE~\cite{lee2022autoregressive}. The technique has found success across various domains, from audio compression in SoundStream~\cite{zeghidour2021soundstream} to vision-language models like DALL-E~\cite{ramesh2021zero} and Stable Diffusion~\cite{rombach2022high}. Recent works such as V2L Tokenizer~\cite{zhu2024beyond} and LLM-AR~\cite{qu2024llms} have demonstrated the potential of quantization for enabling LLMs to process visual and action signals. However, applying these techniques to graphs presents unique challenges due to their irregular structure, unlike images and actions which have natural tokenization patterns through feature maps or temporal sequences.

\begin{figure*}[t]
    \centering
    \includegraphics[width=0.85\textwidth]{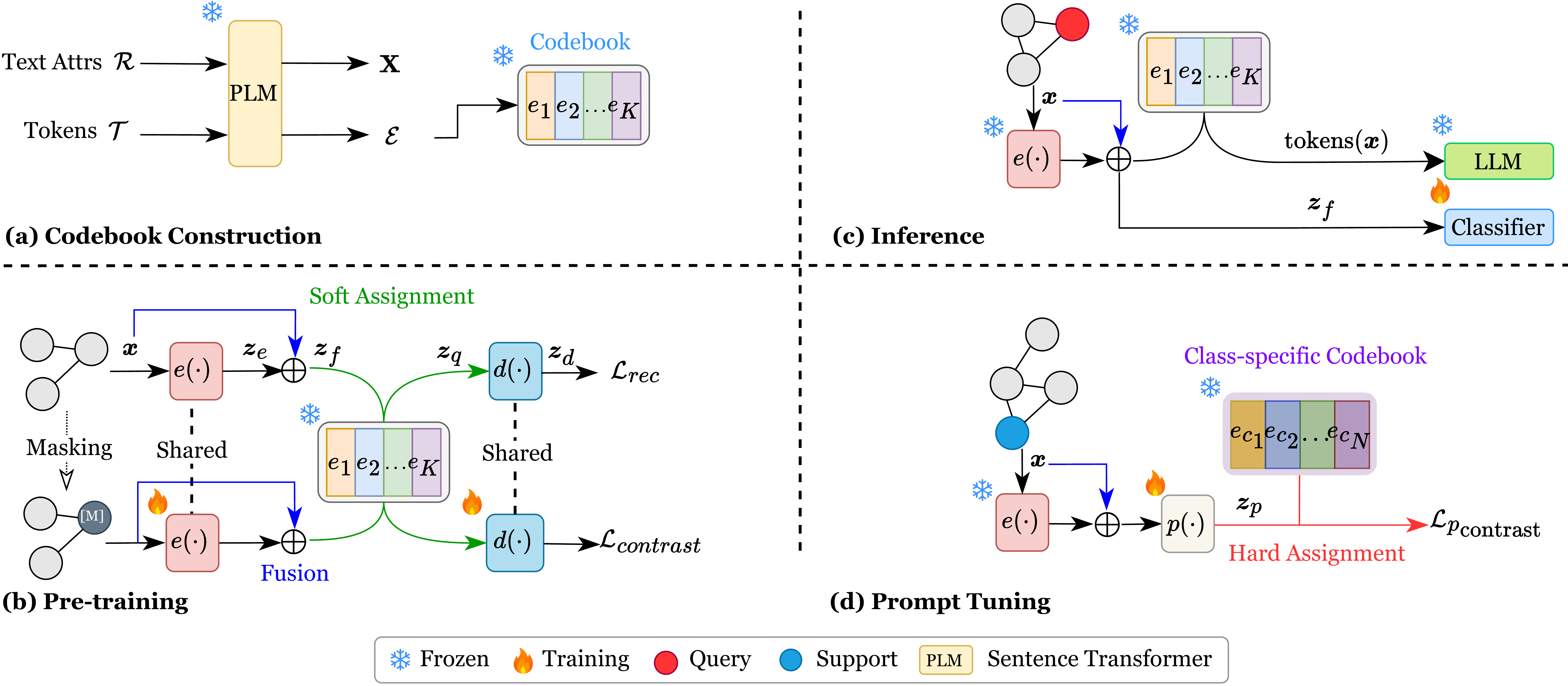}
    \caption{Overview of STAG framework: (a) Feature extraction and codebook construction, (b) Self-supervised pre-training with dual-branch architecture, (c) Inference with or without LLM, and (d) Prompt-tuned inference for few-shot learning.}
    \label{fig:framework}
\end{figure*}

\section{Proposed Approach: STAG}
\label{sec:stag}
We present Soft Tokenization for TAGs (STAG), beginning with the overall framework and then detailing its key stages.

\subsection{Overall Framework}
A text-attributed graph or TAG is defined as $\mathcal{G} = (\mathcal{V}, \mathcal{A}, \mathcal{R})$, where $\mathcal{V}$ is the set of nodes, $\mathcal{A} \in \{0,1\}^{|\mathcal{V}| \times |\mathcal{V}|}$ represents the adjacency matrix, and $\mathcal{R}$ denotes text attributes associated with the nodes. Operating on TAGs, the workflow of STAG consists of three main stages: (1) initial feature extraction and codebook construction, (2) self-supervised pre-training with semantic-structural integration, and (3) flexible inference with or without LLMs. 

As shown in Figure~\ref{fig:framework}(a), we first use a PLM, specifically a sentence transformer~\cite{Reimers2019SentenceBERTSE} to embed raw text attributes into initial node features and construct a codebook from LLM vocabulary. During pre-training in Figure~\ref{fig:framework}(b), we process local subgraphs~\cite{hou2023graphmae2} through a GNN encoder, fuse them with semantic features, and apply soft quantization. The pre-training follows a dual-branch architecture with reconstruction and contrastive objectives. For inference in Figure~\ref{fig:framework}(c), STAG supports both direct classification and LLM-based zero- or few-shot learning. Additionally, prompt tuning in Figure~\ref{fig:framework}(d) further enhances few-shot learning capabilities through a lightweight adaptation mechanism.

\subsection{Initial Feature and Codebook Construction}
VQ-VAE~\cite{van2017neural} provides a foundation for encoding continuous data into discrete representations. Given an input $\boldsymbol{x}$, the encoder $\xi(\cdot)$ maps it to continuous latent vectors $\boldsymbol{z}_e = \xi(\boldsymbol{x})$, which are then quantized by finding their nearest neighbors in a codebook $\mathcal{B} = \{\boldsymbol{e}_k\}_{k=1}^K$ of $K$ embedding vectors: $\boldsymbol{z}_q = \text{Quantize}(\boldsymbol{z}_e) = \boldsymbol{e}_k, \text{ where } k = \arg\min_j \|\boldsymbol{z}_e - \boldsymbol{e}_j\|_2$. The quantized representations $\boldsymbol{z}_q$ are then passed to a decoder $\omega(\cdot)$ to reconstruct the input. 

 Building upon this framework for TAGs, we first encode raw text attributes $\mathcal{R}$ into initial node features $\mathbf{X}\in\mathbb{R}^{|\mathcal{V}| \times d_x}$ using a frozen sentence transformer. For our codebook, we filter LLaMA-2's 32,000 subword vocabulary~\cite{touvron2023llama} by removing non-English and non-alphabetical entries, and eliminating whitespace-based duplicates. The refined token set $\mathcal{T} = \{t_k\}_{k=1}^K$, which has a size of 15,062, is then embedded using the same frozen transformer to create our codebook $\mathcal{E} = \{\boldsymbol{e}_k\}_{k=1}^K$ where $\boldsymbol{e}_k \in \mathbb{R}^{d_x}$, and let $\mathbf{E} = [\boldsymbol{e}_k^\top]_{k=1}^K \in \mathbb{R}^{K \times d_x}$ be the matrix formed by stacking these embeddings.

However, our approach differs from VQ-VAE in several key aspects. First, unlike VQ-VAE's learnable codebook, our codebook $\mathcal{E}$ remains frozen throughout pre-training to ensure semantic consistency. Second, instead of hard nearest-neighbor quantization, we introduce soft assignment to better handle the lack of natural tokenization structures in graphs. Third, we incorporate semantic-structural fusion and distribution alignment mechanisms to effectively preserve both types of information. These innovations, detailed in the following sections, enable effective graph representation learning while maintaining LLM compatibility.

\subsection{Pre-training with Soft Assignment}
To address the challenges of integrating both semantic and structural information into LLM-compatible formats, we develop a self-supervised pre-training approach with feature fusion and soft assignment strategy, ensuring consistency across different LLMs.

\subsubsection{Structural-Semantic Feature Fusion}
Following GraphMAE's masking strategy~\cite{hou2022graphmae}, we randomly mask nodes $\widetilde{\mathcal{V}} \subset \mathcal{V}$ with a learnable [M] token while keeping the graph structure intact. Our framework processes both original and masked graphs through parallel branches sharing the same feature learning and quantization mechanisms, employing a GNN encoder $e(\cdot)$ to learn structural representations $\boldsymbol{z}_e = e(\boldsymbol{x}) \in \mathbb{R}^{d_h}$, where $\boldsymbol{x}$ is the node feature vector from $\mathbf{X}$. However, directly quantizing these structural embeddings risks losing the rich semantic information present in the original text attributes.

Therefore, to preserve both structural and semantic information while maintaining computational efficiency, we introduce a parameter-efficient feature fusion module. Given the GNN output $\boldsymbol{z}_e$, we first project it to dimension $d_x$ (matching the initial feature dimension) using a linear projection layer $\mathbf{W}_f \in \mathbb{R}^{d_x \times d_h}$. Then, combined with initial node features $\boldsymbol{x}$, we compute:
\begin{equation}
    \boldsymbol{z}_f = \phi \cdot \frac{\mathbf{W}_f\boldsymbol{z}_e}{\|\mathbf{W}_f\boldsymbol{z}_e\|_2} + \psi \cdot \frac{\boldsymbol{x}}{\|\boldsymbol{x}\|_2},
\end{equation}
where $\phi$ and $\psi \in (0, 1]$ are learnable parameters. This design is motivated by two key insights: (1) both features originate from the same text input, allowing for efficient integration without complex transformations, and (2) L2 normalization preserves the directional information of both structural and semantic features.

\subsubsection{Soft Assignment with Distribution Alignment}
The fused features $\boldsymbol{z}_f \in \mathbb{R}^{d_x}$ are then quantized into the discrete token space through our semantic-preserving quantization process. 

\stitle{Soft Assignment Strategy.} We develop a soft assignment strategy that maps each node to a distribution of tokens, addressing two key challenges: (1) the absence of explicit tokenization structures in graphs, and (2) preventing overfitting to specific tokens to improve transferability across domains. Unlike VQ-VAE, which relies on L2 distance and hard assignment for codebook lookup, we leverage cosine similarity to compute attention weights over the entire codebook, enabling soft token assignments for each fused node representation $\boldsymbol{z}_f$:
\begin{equation}
  \text{attn}(\boldsymbol{z}_f) = \text{softmax}\left(\left[\theta(\boldsymbol{z}_f, \boldsymbol{e}_k)\right]_{k=1}^K/{\tau_{sa}}\right),
  \label{eq:attn}
\end{equation}
where $\theta(\boldsymbol{z}_f, \boldsymbol{e}_k)$ computes the cosine similarity between $\boldsymbol{z}_f$ and the $k$-th codebook embedding, and $\tau_{sa}$ is a temperature parameter that controls the softness of the assignment. This cosine similarity-based attention is particularly suitable for our semantic codebook space, as it focuses on directional similarity rather than absolute distances. The quantized representation is then computed as the weighted sum of the codebook embeddings:
\begin{equation}
    \boldsymbol{z}_q = \mathbf{E}^\top \text{attn}(\boldsymbol{z}_f).
    \label{eq:soft}
\end{equation}
To prevent encoder outputs from diverging too far from the codebook space during training, we employ a commitment loss:
\begin{equation}
  \mathcal{L}_{\text{commit}} = \beta\left(1 - \theta(\boldsymbol{z}_f, \text{sg}[\boldsymbol{z}_q])\right),
\end{equation}
where sg[·] denotes the stop-gradient operator and $\beta \in (0, 2]$ controls the strength of the commitment. While VQ-VAE uses L2-based hard assignment for both lookup and commitment loss, we adopt cosine similarity for semantic space alignment and soft assignments through weighted codebook combinations. Although the frozen codebook ensures consistent semantics across LLMs, the soft assignment alone cannot guarantee that the quantized representations preserve the semantic meaning of the original node text.

\stitle{Self-supervised Semantic Alignment.} To address this challenge, we incorporate a KL divergence loss that aligns the attention distribution of fused features with that of the original semantic features to enhance the alignment between structural and semantic representations in the absence of labeled data:
\begin{equation}
    \mathcal{L}_{\text{KL}} = D_{\text{KL}}(\text{attn}(\boldsymbol{x}) \| \text{attn}(\boldsymbol{z}_f)),
\end{equation}
where $\text{attn}(\boldsymbol{x})$ represents the attention distribution computed from the original node features $\boldsymbol{x}$ over the codebook following Eq.~\eqref{eq:attn}. 

The KL divergence aligns token assignment distributions between fused and original features, enabling effective integration of structural and semantic information by guiding the structure-semantic fused representations to be quantized to semantically meaningful tokens. Unlike Gumbel-Softmax~\cite{jang2016categorical}, which uses KL divergence to train discrete categorical distributions, our KL loss guides the soft assignment to maintain semantic consistency with respect to the frozen codebook without requiring labels.

\subsubsection{Dual-Branch Training Objectives}
Our pre-training maintains parallel branches sharing the same quantization mechanism while serving different learning objectives: reconstruction for node-level semantics and contrastive learning for neighborhood structures.

\stitle{Reconstruction Branch.} This branch processes the original graph to preserve node-level semantic information through reconstruction. Given a shared GNN decoder $d(\cdot)$ that maps quantized representations back to the initial feature dimension $d_x$, we extend the traditional VQ-VAE reconstruction to graphs using GraphMAE's scaled cosine error (SCE) loss:
\begin{equation}
    \mathcal{L}_{\text{rec}} = \frac{1}{|\mathcal{V}|} \sum_{v_i \in \mathcal{V}} \left(1 - \frac{\boldsymbol{x}_i^\top \boldsymbol{z}_{d_i}}{\|\boldsymbol{x}_i\| \cdot \|\boldsymbol{z}_{d_i}\|}\right)^\gamma,
\end{equation}
where $\boldsymbol{z}_d=d(\boldsymbol{z}_q) \in \mathbb{R}^{d_x}$ is the decoded feature of the quantized representation, and $\gamma \geq 1$ is a scaling factor that controls the penalty for reconstruction errors. Unlike traditional mean squared error used in VQ-VAE, SCE loss avoids sensitivity and low selectivity issues in graph representation learning~\cite{hou2023graphmae2}, making it particularly suitable for preserving semantic information during reconstruction.

\stitle{Contrastive Branch.} This branch processes the masked graph to capture neighborhood structural patterns through a contrastive learning objective. For each masked node $v_i \in \widetilde{\mathcal{V}}$, we form positive pairs between its decoded features $\boldsymbol{z}_{d_i}$ and original features $\boldsymbol{x}_i$, with other masked nodes' decoded features serving as negatives. Following~\cite{oord2018representation}, the contrastive loss is defined as:
\begin{equation}
\small
\begin{split}
\ell\left(\boldsymbol{z}_{d_i}, \boldsymbol{x}_i\right)=\log \Bigg(e^{\theta\left(\boldsymbol{z}_{d_i}, \boldsymbol{x}_i\right)/ \tau_c} / \Bigg(\underbrace{e^{\theta\left(\boldsymbol{z}_{d_i}, \boldsymbol{x}_i\right) / \tau_c}}_{\text {positive pair}}+ \\
\underbrace{\sum_{j=1,j \neq i}^{|\mathcal{N}_i|} e^{\theta\left(\boldsymbol{z}_{d_i}, \boldsymbol{z}_{d_j}\right) / \tau_c}}_{\text{negative pairs}}\Bigg)\Bigg), \forall v_i \in \widetilde{\mathcal{V}} \text{ and } \mathcal{N}_i \subseteq \widetilde{\mathcal{V}},
\end{split}
\end{equation}
where $\mathcal{N}_i$ represents a subset of masked nodes selected as negative samples for $v_i$, $\tau_c$ is the temperature parameter. The overall contrastive loss is $\mathcal{L}_{\text{contrast}} = -\frac{1}{|\widetilde{\mathcal{V}}|}\sum_{v_i \in \widetilde{\mathcal{V}}}\ell(\boldsymbol{z}_{d_i}, \boldsymbol{x}_i)$.
This contrastive objective compels the model to utilize neighborhood structural information for feature prediction~\cite{xie2022self}, capturing essential graph topology in quantized representations. 
The final training objective combines losses from both branches:
\begin{equation}
  \mathcal{L} = \underbrace{\frac{1}{2}(\mathcal{L}_{\text{commit}}^{\text{Rec}} + \mathcal{L}_{\text{commit}}^{\text{Contrast}})}_{\text{commitment}} + \underbrace{\mathcal{L}_{\text{rec}}}_{\text{semantic}} + \underbrace{\mathcal{L}_{\text{contrast}}}_{\text{structural}} + \underbrace{\lambda\mathcal{L}_{\text{KL}}}_{\text{alignment}},
\end{equation}

where $\lambda$ is a balancing parameter, and $\mathcal{L}_{\text{commit}}^{\text{Rec}}$ and $\mathcal{L}_{\text{commit}}^{\text{Contrast}}$ are commitment losses applied to reconstruction and contrastive branches, respectively. This multi-objective optimization ensures effective integration of semantic and structural information while ensuring compatibility with discrete token spaces.

\subsection{Flexible Inference with or without LLMs}
\label{sec:inference}
Through effective semantic-structural quantization, our framework bridges continuous graph embeddings and discrete token spaces, enabling both LLM-based and traditional inference strategies.

\stitle{Inference with LLMs.} To convert continuous node embeddings into discrete tokens for LLM processing, we first process each node with its local subgraph through our pre-trained GNN encoder and fusion module to obtain its learned embedding. We then compute its attention distribution over the codebook $\mathcal{E}$ and select the $top\text{-}k \in \mathbb{N}^+$ corresponding tokens from $\mathcal{T}$ with the highest attention weights:
\begin{equation}
  \begin{split}
      & \text{tokens}(\boldsymbol{x}) = [t_{m_1}, t_{m_2}, ..., t_{m_{top\text{-}k}}], \text{ where } \\
      & m_i = \underset{j \in \{1,...,K\} \setminus \{m_1,...,m_{i-1}\}}{\text{argmax }} \text{attn}(\boldsymbol{z}_f), \text{ for } i \in [1,top\text{-}k].
  \end{split}
\end{equation}
For $N$-way $k$-shot few-shot learning where $N$ represents the number of classes and $k$ is the number of examples per class, we construct a system prompt that includes $N \times k$ support examples, where each example consists of a node's tokens and its corresponding class label. The LLM then predicts the class of a new test node based on its tokens and these support examples. Below is a simplified example using a 3-way 1-shot setting:
\begin{tcolorbox}[boxsep=0mm,left=2.5mm,right=2.5mm]
\footnotesize
\textbf{System Prompt}: You are a node classifier. Given a list of tokens representing a node's features, predict its class from the following options: [Research Paper, Dataset, Software].

\textbf{Few-shot Examples}:
Node tokens: [research, methodology, experiment]
Class: Research Paper

Node tokens: [benchmark, statistics, collection]
Class: Dataset

Node tokens: [implementation, code, library]
Class: Software

\textbf{Test Node}: Node tokens: [algorithm, computation, optimization]
Predict the class:
\end{tcolorbox}

Similarly, for zero-shot learning, we directly query the LLM with only the system prompt and test node tokens. Below is a simplified 4-way zero-shot example (see {Appendix~\ref{sec:appendix_prompt}} for the complete prompt templates for both few-shot and zero-shot):

\begin{tcolorbox}[boxsep=0mm,left=2.5mm,right=2.5mm]
\footnotesize
\textbf{System Prompt}: You are a node classifier. Given a list of tokens representing a node's features, predict its class from the following options: [Research Paper, Dataset, Software, Survey Paper].

\textbf{Test Node}:

Node tokens: [algorithm, computation, optimization]
Predict the class:
\end{tcolorbox}
Our quantized tokens integrate both semantic and structural information from input nodes, enabling us to leverage different LLM capabilities effectively. In few-shot learning, we combine the LLM's in-context learning ability (through support examples) with its semantic understanding to make predictions. For zero-shot learning, we rely purely on the LLM's semantic knowledge, allowing classification without any labeled examples.

\stitle{Inference without LLMs.} For traditional inference, following standard practice in graph self-supervised learning~\cite{velickovic2019deep,hassani2020contrastive,zhang2021canonical,thakoor2022large}, we employ linear probing to evaluate the quality of learned representations. Specifically, we freeze the pre-trained GNN encoder and fusion module, and train only a linear classifier on the fused features $\boldsymbol{z}_f$. Given a linear classifier $\mathbf{W}_c \in \mathbb{R}^{N \times d_x}$ where $N$ is the number of candidate classes, the prediction is computed as:
\begin{equation}
    \hat{y} = \text{argmax}(\text{softmax}(\mathbf{W}_c\boldsymbol{z}_f)).
\end{equation}
This approach allows direct comparison with other graph SSL methods while maintaining the efficiency of our framework. Beyond these basic inference strategies, our framework enables enhanced few-shot domain transfer through a prompt tuning mechanism that can operate both with and without LLMs.

\subsection{Prompt-tuning for Domain Transfer}
To enhance domain transfer capabilities in few-shot settings, we develop a prompt tuning mechanism~\cite{wen2024prompt} that builds on our quantization framework while switching to hard token assignment. During prompt tuning, we keep our pre-trained GNN encoder and fusion module frozen, and introduce a lightweight prompt network $p(\cdot)$ (two-layer bottleneck neural network) that processes the fused features. Specifically, the prompted features $\boldsymbol{z}_p \in \mathbb{R}^{d_x}$ are computed as: $\boldsymbol{z}_p = p(\boldsymbol{z}_f) \odot \boldsymbol{z}_f$, where $\odot$ denotes element-wise multiplication. For $N$-way $k$-shot setting, we obtain class label explanations via LLMs following \cite{he2023harnessing,liu2023one} and embed them using the same frozen sentence transformer to create a frozen class-specific codebook $\mathcal{C} = \{\boldsymbol{e}_{c_n}\}_{n=1}^N \in \mathbb{R}^{N \times d_x}$, where $\boldsymbol{e}_{c_n}$ represents the embedded explanation of class $c_n$.

The prompt tuning is guided by two objectives. First, following our quantization design, a commitment loss ensures the prompted embeddings remain close to $\mathcal{C}$:
\begin{equation}
    \mathcal{L}_{p_\text{commit}} = \beta_p\left(1 - \theta(\boldsymbol{z}_p, \text{sg}[\boldsymbol{z}_q])\right),
\end{equation}
where $\beta_p$ controls the strength of the commitment.

Second, with $N\times k$ labeled support examples, we employ a weighted contrastive loss that leverages class labels to guide the tuning:
\begin{equation}
    \mathcal{L}_{p_\text{contrast}} = -\frac{1}{Nk} \sum_{i=1}^{Nk} \sum_{j=1}^N \theta(c(\boldsymbol{x}_i),c_j) \log \frac{e^{\theta(\boldsymbol{z}_{p_i}, \boldsymbol{e}_{c_j})/\tau_p}}{\sum_{n=1}^N e^{\theta(\boldsymbol{z}_{p_i}, \boldsymbol{e}_{c_n})/\tau_p}},
\end{equation}
where $\theta(c(\boldsymbol{x}_i),c_j)$ represents the semantic similarity between the ground truth class $c(\boldsymbol{x}_i)$ of example $\boldsymbol{x}_i$ and class $c_j$. This weighted objective encourages the prompted embeddings to be similar to their ground truth class embeddings while considering semantic relationships between classes. This design helps capture the natural hierarchy and relationships between classes, leading to more robust few-shot learning.

For inference with LLMs, we quantize the prompted embeddings using our original codebook $\mathcal{E}$ to obtain tokens (we use the same system prompt as in Section \ref{sec:inference}). For inference without LLMs, we directly compute similarity with the class embeddings, where the most similar class embedding indicates the predicted class:
\begin{equation}
  \hat{y} = \underset{n \in \{1,...,N\}}{\text{argmax }} \theta(\boldsymbol{z}_p, \boldsymbol{e}_{c_n}).
  \label{eq:c_prediction}
\end{equation}
This unified prompt tuning approach enables effective domain transfer by leveraging class semantics while maintaining compatibility with both LLM-based and traditional classification paths.

\begin{table*}[t]
  \centering
  \small
  \caption{5-way 5-shot node classification across different datasets (except PubMed: 3-way). Models are pre-trained on Cora Full or ogbn-products with $top\text{-}k=13$ for LLM inference. Gray-shaded rows: supervised baselines trained directly on target datasets; Colored cells: pre-train dataset matches target (blue: Cora Full, green: ogbn-products). Results show accuracy (\%) averaged over 20 random tasks, with best results among our variants in bold.}
  \begin{threeparttable}
  \renewcommand{\arraystretch}{0.95}
  \resizebox{0.95\linewidth}{!}{
  \begin{tabular}{l|c|c|ccccccc}
      \toprule[1.2pt]
      \multirow{2}{*}{Pre-train data} & \multirow{2}{*}{Method} & \multirow{2}{*}{LLM} & \multicolumn{7}{c}{Target data}\\
      \cline{4-10}
       &   &   & Cora & Cora Full & CiteSeer & PubMed & WikiCS & ogbn-arxiv & ogbn-products \\
       \midrule \midrule
       \multirow{2}{*}{Same as target} & \cellcolor{lightgray} GCN & \cellcolor{lightgray} \ding{55} & \cellcolor{lightgray} 76.10{\scriptsize$\pm$4.26} & \cellcolor{lightgray} 82.81{\scriptsize$\pm$7.40} & \cellcolor{lightgray} 59.95{\scriptsize$\pm$6.92} & \cellcolor{lightgray} 66.35{\scriptsize$\pm$6.71} & \cellcolor{lightgray} 70.55{\scriptsize$\pm$8.26} & \cellcolor{lightgray} 76.61{\scriptsize$\pm$7.72} & \cellcolor{lightgray} 80.08{\scriptsize$\pm$7.41} \\
       & \cellcolor{lightgray} GAT & \cellcolor{lightgray} \ding{55} & \cellcolor{lightgray} 79.60{\scriptsize$\pm$5.00} & \cellcolor{lightgray} 84.72{\scriptsize$\pm$7.83} & \cellcolor{lightgray} 60.85{\scriptsize$\pm$6.78} & \cellcolor{lightgray} 67.40{\scriptsize$\pm$7.18} & \cellcolor{lightgray} 77.95{\scriptsize$\pm$7.81} & \cellcolor{lightgray} 80.80{\scriptsize$\pm$7.99} & \cellcolor{lightgray} 81.94{\scriptsize$\pm$7.07} \\
       \midrule
    \multirow{3}{*}{No pre-train} & Raw Text & \ding{51} & 63.40{\scriptsize$\pm$9.07} & 71.66{\scriptsize$\pm$7.84} & 62.10{\scriptsize$\pm$5.35} & 85.00{\scriptsize$\pm$5.48} & 77.15{\scriptsize$\pm$6.92} & 54.74{\scriptsize$\pm$9.21} & 87.58{\scriptsize$\pm5.48$} \\
      & Raw Feat + Quantization & \ding{51} & 54.85{\scriptsize$\pm$6.28} & 73.74{\scriptsize$\pm$8.10} & 56.40{\scriptsize$\pm$5.48} & 48.30{\scriptsize$\pm$7.78} & 73.40{\scriptsize$\pm$8.19} & 63.75{\scriptsize$\pm$9.98} & 65.84{\scriptsize$\pm$9.96} \\
       & Raw Feat + Linear Probing & \ding{55} & 70.25{\scriptsize$\pm$7.22} & 81.29{\scriptsize$\pm$7.47} & 63.00{\scriptsize$\pm$6.72} & 68.30{\scriptsize$\pm$6.28} & 78.05{\scriptsize$\pm$7.47} & 83.05{\scriptsize$\pm$7.40} & 77.53{\scriptsize$\pm$7.16} \\
      \midrule[0.8pt]
      \multirow{11}{*}{Cora Full} 
      & DGI & \ding{55} & 77.05{\scriptsize$\pm$5.12} & \cellcolor{blue!20}83.32{\scriptsize$\pm$8.12} & 63.85{\scriptsize$\pm$5.39} & 68.20{\scriptsize$\pm$7.57} & 78.65{\scriptsize$\pm$6.90} & 81.30{\scriptsize$\pm$8.51} & 79.90{\scriptsize$\pm$7.20} \\
      & GraphMAE2 & \ding{55} & 77.70{\scriptsize$\pm$6.92} & \cellcolor{blue!20}84.74{\scriptsize$\pm$7.42} & 65.25{\scriptsize$\pm$5.84} & 66.35{\scriptsize$\pm$6.09} & 80.95{\scriptsize$\pm$4.96} & 80.04{\scriptsize$\pm$8.15} & 73.93{\scriptsize$\pm$7.57} \\ \noalign{\vskip -2pt}  
        \cmidrule{2-10}
        \noalign{\vskip -2pt}  
      & GPPT & \ding{55} & 27.16{\scriptsize$\pm$7.61} & \cellcolor{blue!20}67.90{\scriptsize$\pm$12.72} & 28.66{\scriptsize$\pm$7.60} & 21.53{\scriptsize$\pm$10.91} & 29.00{\scriptsize$\pm$8.08} & 36.92{\scriptsize$\pm$10.32} & 24.32{\scriptsize$\pm$5.13} \\
      & G2P2 & \ding{55} & 74.90{\scriptsize$\pm$7.47} & \cellcolor{blue!20}81.10{\scriptsize$\pm$7.44} & 59.65{\scriptsize$\pm$9.68} & 67.85{\scriptsize$\pm$8.02} & 69.90{\scriptsize$\pm$10.52} & 68.75{\scriptsize$\pm$10.14} & 70.97{\scriptsize$\pm$10.03} \\
        \noalign{\vskip -2pt}  
        \cmidrule{2-10}
        \noalign{\vskip -2pt}  
      & Prodigy & \ding{55} & 39.50{\scriptsize$\pm$6.75} & \cellcolor{blue!20}60.80{\scriptsize$\pm$6.38} & 42.90{\scriptsize$\pm$5.02} & 43.68{\scriptsize$\pm$6.91} & 43.25{\scriptsize$\pm$6.91} & 47.85{\scriptsize$\pm$6.89} & 30.70{\scriptsize$\pm$5.94} \\
      & OFA & \ding{55} & 45.95{\scriptsize$\pm$4.52} & \cellcolor{blue!20}56.95{\scriptsize$\pm$5.31} & 36.80{\scriptsize$\pm$5.50} & 49.40{\scriptsize$\pm$4.75} & 46.45{\scriptsize$\pm$4.67} & 50.80{\scriptsize$\pm$4.73} & 33.60{\scriptsize$\pm$4.26} \\
        \noalign{\vskip -2pt}  
        \cmidrule{2-10}
        \noalign{\vskip -2pt}  
      & STAG & \ding{51} & 67.60{\scriptsize$\pm$6.72} & \cellcolor{blue!20}80.95{\scriptsize$\pm$8.02} & 62.45{\scriptsize$\pm$7.02} & 54.50{\scriptsize$\pm$7.83} & 79.20{\scriptsize$\pm$8.41} & 71.56{\scriptsize$\pm$10.32} & 69.34{\scriptsize$\pm$9.93} \\
      & + Linear Probing & \ding{55} & 78.50{\scriptsize$\pm$5.62} & \cellcolor{blue!20}86.04{\scriptsize$\pm$6.70} & \bf 66.70{\scriptsize$\pm$5.36} & \bf 69.00{\scriptsize$\pm$6.31} & \bf 84.05{\scriptsize$\pm$5.78} & 82.99{\scriptsize$\pm$8.10} & 79.62{\scriptsize$\pm$7.12} \\
      & + Prompt Tuning & \ding{51} & 73.30{\scriptsize$\pm$4.77} & \cellcolor{blue!20}85.20{\scriptsize$\pm$7.59} & 65.40{\scriptsize$\pm$5.98} & 66.20{\scriptsize$\pm$5.70} & 79.45{\scriptsize$\pm$7.53} & 79.18{\scriptsize$\pm$8.28} & 73.94{\scriptsize$\pm$9.67} \\
      & + Prompt Tuning* & \ding{55} & \bf 78.65{\scriptsize$\pm$5.93} & \cellcolor{blue!20}\bf 86.66{\scriptsize$\pm$7.67} &
      65.80{\scriptsize$\pm$7.03} & 68.25{\scriptsize$\pm$6.80} & 83.55{\scriptsize$\pm$5.94} & \bf 83.57{\scriptsize$\pm$8.30} & \bf 80.48{\scriptsize$\pm$6.86} \\
      \midrule[0.8pt]
      \multirow{11}{*}{ogbn-products}
      & DGI & \ding{55} & OOM & OOM & OOM & OOM & OOM & OOM & \cellcolor{green!20}OOM \\
      & GraphMAE2 & \ding{55} & 69.30{\scriptsize$\pm$5.51} & 77.74{\scriptsize$\pm$7.53} & 55.90{\scriptsize$\pm$7.73} & 60.05{\scriptsize$\pm$6.05} & 71.15{\scriptsize$\pm$7.68} & 67.70{\scriptsize$\pm$9.51} & \cellcolor{green!20}81.08{\scriptsize$\pm$8.00} \\ \noalign{\vskip -2pt}  
        \cmidrule{2-10}
        \noalign{\vskip -2pt}  
      & GPPT & \ding{55} & 24.52{\scriptsize$\pm$4.77} & 25.30{\scriptsize$\pm$6.61} & 23.76{\scriptsize$\pm$4.70} & 24.70{\scriptsize$\pm$10.84} & 23.32{\scriptsize$\pm$5.42} & 26.52{\scriptsize$\pm$6.49} & \cellcolor{green!20}55.18{\scriptsize$\pm$11.89} \\
      & G2P2 & \ding{55} & 70.70{\scriptsize$\pm$9.00} & 76.35{\scriptsize$\pm$8.62} & 53.10{\scriptsize$\pm$9.97} & 64.95{\scriptsize$\pm$9.40} & 69.20{\scriptsize$\pm$9.51} & 65.55{\scriptsize$\pm$9.07} & \cellcolor{green!20}75.52{\scriptsize$\pm$10.10} \\ \noalign{\vskip -2pt}  
        \cmidrule{2-10}
        \noalign{\vskip -2pt}  
      & Prodigy & \ding{55} & 28.90{\scriptsize$\pm$5.38} & 36.54{\scriptsize$\pm$6.04} & 28.00{\scriptsize$\pm$4.56} & 47.11{\scriptsize$\pm$7.68} & 33.85{\scriptsize$\pm$5.82} & 36.65{\scriptsize$\pm$5.17} & \cellcolor{green!20}62.20{\scriptsize$\pm$7.29} \\
      & OFA & \ding{55} & 25.85{\scriptsize$\pm$3.12} & 29.70{\scriptsize$\pm$5.09} & 23.05{\scriptsize$\pm$3.81} & 51.30{\scriptsize$\pm$5.53} & 33.80{\scriptsize$\pm$4.76} & 28.95{\scriptsize$\pm$4.07} & \cellcolor{green!20}59.65{\scriptsize$\pm$5.70} \\ \noalign{\vskip -2pt}  
        \cmidrule{2-10}
        \noalign{\vskip -2pt}  
      & STAG & \ding{51} & 59.20{\scriptsize$\pm$7.07} & 73.60{\scriptsize$\pm$8.00} & 54.90{\scriptsize$\pm$6.11} & 48.45{\scriptsize$\pm$7.61} & 76.15{\scriptsize$\pm$8.09} & 68.56{\scriptsize$\pm$10.30} & \cellcolor{green!20}79.75{\scriptsize$\pm$9.33} \\
      & + Linear Probing & \ding{55} & 73.65{\scriptsize$\pm$7.49} & 83.14{\scriptsize$\pm$6.26} & 62.30{\scriptsize$\pm$6.08} & \bf 69.85{\scriptsize$\pm$5.89} & 80.35{\scriptsize$\pm$7.17} & \bf 81.62{\scriptsize$\pm$7.26} & \cellcolor{green!20} 83.94{\scriptsize$\pm$6.78} \\
      & + Prompt Tuning & \ding{51} & 68.95{\scriptsize$\pm$5.93} & 80.07{\scriptsize$\pm$7.82} & 60.40{\scriptsize$\pm$6.63} & 63.70{\scriptsize$\pm$5.72} & 78.55{\scriptsize$\pm$8.24} & 74.85{\scriptsize$\pm$9.89} & \cellcolor{green!20}82.35{\scriptsize$\pm$6.24} \\
      & + Prompt Tuning* & \ding{55} & \bf 74.90{\scriptsize$\pm$6.54} & \bf 83.52{\scriptsize$\pm$7.60} & \bf 63.70{\scriptsize$\pm$6.00} & 68.45{\scriptsize$\pm$6.20} & \bf 81.40{\scriptsize$\pm$7.14} & 81.19{\scriptsize$\pm$7.68} & \cellcolor{green!20}\bf 84.03{\scriptsize$\pm$5.54} \\
      \bottomrule[1.2pt]
      \end{tabular}}
      \end{threeparttable}
      \begin{tablenotes}
      \footnotesize
      \item \ding{51}/\ding{55}: LLM usage during inference; \textbf{Prompt Tuning*}: inference without LLM; \textbf{Raw Text}: Use raw text for LLM inference (To fit in the context window of LLM, raw text is truncated); \textbf{Raw Feat + Quantization}: Directly quantize raw node features into tokens for LLM inference; \textbf{Raw Feat + Linear Probing}: Train a linear classifier on raw node features without any pre-training; \textbf{OOM}: Out-of-memory error during training.
      \end{tablenotes}
      \label{tab:few_shot}
\end{table*}

\section{Experiments}
We conduct comprehensive experiments to evaluate our framework's ability to bridge graph representation learning with LLMs. Specifically, we investigate three key research questions:
\begin{itemize}
    \item \textbf{RQ1}: How effectively does STAG integrate structural and semantic information in few-shot learning?
    \item \textbf{RQ2}: Can our framework enable effective zero-shot generalization across different domains?
    \item \textbf{RQ3}: How flexible is our framework in supporting different LLM architectures?
\end{itemize}

\subsection{Experimental Setup}
\label{sec:exp_setup}
We evaluate our framework on seven text-attributed graph datasets: five citation networks (Cora Full~\cite{mccallum2000automating}, its pruned subset standard Cora, CiteSeer, PubMed~\cite{yang2016revisiting}, ogbn-arxiv~\cite{hu2020open}), one web graph (WikiCS~\cite{mernyei2020wiki}), and one co-purchase network (a subset of ogbn-products~\cite{hu2020open} following~\cite{he2023harnessing}). We use sentence transformers~\cite{Reimers2019SentenceBERTSE} to obtain $768$-dimensional embeddings as initial node features. For our method, this same model is also used for codebook construction. And we use GAT as both encoder and decoder in our framework.

We compare with three categories of baselines: (1) Traditional graph learning methods including GCN~\cite{thomas2017gcn} and GAT~\cite{velivckovic2018graph}, and self-supervised approaches like DGI~\cite{velickovic2019deep} and GraphMAE2~\cite{hou2023graphmae2}, which serve as strong baselines for representation learning on graphs; (2) Few-shot graph learning methods including GPPT~\cite{sun2022gppt} and G2P2~\cite{wen2024prompt}, which are specifically designed for few-shot learning on graphs through prompt tuning; (3) GraphLLM methods like Prodigy~\cite{huang2024prodigy} and OFA~\cite{liu2023one}, which both require labeled data from source datasets during training for model adaptation.

During pre-training, OFA and Prodigy require labeled data with specific class-splitting strategies, while all other methods use unlabeled data. For ogbn-arxiv, we follow OFA's splitting strategy; for Cora Full and ogbn-products, we create splits with non-overlapping classes. For evaluation, we conduct 5-way 5-shot experiments across 20 random tasks, with a total of 2000 balanced queries distributed across all tasks. For OFA and Prodigy, tasks are created from their test splits (classes unseen during training) to ensure fair evaluation of their transfer learning capabilities. For all other methods that don't require labeled pre-training data, tasks are created from the full dataset. For LLM inference, we use $top\text{-}k=13$ tokens based on empirical performance. All experiments use LLaMA3-8B~\cite{dubey2024llama} by default and report classification accuracy. Detailed hyperparameter configurations and hardware specifications are provided in Appendix~\ref{sec:appendix_implementation}.

\subsection{RQ1: Cross-dataset Few-shot Learning}
\label{sec:few_shot}
Table~\ref{tab:few_shot} presents the cross-dataset few-shot learning results across different source-target dataset combinations, where all methods are pre-trained on either Cora Full or ogbn-products except GCN and GAT. Our framework demonstrates three key advantages. (1) \emph{Effective structural-semantic integration}: Our framework achieves competitive results compared to traditional GNNs and self-supervised methods like GraphMAE2, validating our learned representations. Unlike G2P2 which uses dataset-specific text-graph alignment with a trainable text encoder, STAG prioritizes generalization through a frozen sentence transformer with parameter-efficient alignment modules, showing robust performance while outperforming OFA despite its need for source dataset labels. (2) \emph{Framework versatility}: Prompt tuning enhances performance in both inference paths - with LLM (`STAG+Prompt Tuning') achieving competitive results, and without LLM (`STAG+Prompt Tuning*') achieving state-of-the-art performance on several datasets. (3) \emph{Robust transfer ability}: When pre-trained on ogbn-products, Prodigy and OFA suffer significant performance drops when tested on different target datasets. In contrast, our framework maintains strong performance on citation networks like ogbn-arxiv, validating its ability to capture generalizable graph representations despite domain shifts. When compared to raw text and feature variants, our method outperforms both `Raw Feat + Quantization' and `Raw Text' with LLM by effectively incorporating structural information into learned tokens for both inference paths, particularly in Cora Full and ogbn-arxiv datasets despite raw texts' potential information leakage. Results for models pre-trained on ogbn-arxiv are in Appendix~\ref{sec:appendix_exp}.

\begin{table}[t]
    \centering
    \caption{5-way zero-shot classification results ($top\text{-}k=13$).}
    \begin{threeparttable}
    \setlength{\tabcolsep}{0.9pt}
    \resizebox{\linewidth}{!}{
    \begin{tabular}{l|c|c|cccc}
        \toprule[1.2pt]
        \multirow{2}{*}{Pre-train data} & \multirow{2}{*}{Method} & \multirow{2}{*}{LLM} & \multicolumn{4}{c}{Target data}\\\cline{4-7}
        &   &   & Cora & Cora Full & WikiCS & ogbn-arxiv \\
        \midrule \midrule
        \multirow{2}{*}{No pre-train}& Raw Feat + Q& \ding{51} & 47.10{\scriptsize$\pm$5.98} & 60.33{\scriptsize$\pm$10.88} & 70.40{\scriptsize$\pm$8.88} & 25.48{\scriptsize$\pm$5.54} \\
        & Raw Feat + $\mathcal{C}$ & \ding{55} & 62.20{\scriptsize$\pm$8.45} & 77.23{\scriptsize$\pm$8.96} & 73.85{\scriptsize$\pm$8.02} & 72.85{\scriptsize$\pm$10.43}\\
        \midrule
        \multirow{4}{*}{Cora Full} 
        & G2P2 & \ding{55} & 60.45{\scriptsize$\pm$7.58} & \cellcolor{blue!20}64.29{\scriptsize$\pm$11.56} &  50.25{\scriptsize$\pm$8.43} & 19.66{\scriptsize$\pm$6.38}\\
        & OFA & \ding{55} & 20.30{\scriptsize$\pm$2.93} & \cellcolor{blue!20}23.85{\scriptsize$\pm$3.58}  & 21.45{\scriptsize$\pm$3.99} & 17.60{\scriptsize$\pm$3.74} \\
        \noalign{\vskip -2pt}  
        \cmidrule{2-7}
        \noalign{\vskip -2pt}  
        & STAG & \ding{51} & 48.05{\scriptsize$\pm$6.15} & \cellcolor{blue!20}62.63{\scriptsize$\pm$11.70} & \bf 76.25{\scriptsize$\pm$8.48} & 26.01{\scriptsize$\pm$7.52} \\
        & STAG + $\mathcal{C}$ & \ding{55} & \bf 66.55{\scriptsize$\pm$7.48} & \cellcolor{blue!20}\bf 82.90{\scriptsize$\pm$9.52}  & 75.15{\scriptsize$\pm$7.81} & \bf 74.23{\scriptsize$\pm$9.35} \\
        \bottomrule[1.2pt]
    \end{tabular}}
    \end{threeparttable}
    \begin{tablenotes}
    \footnotesize
    \item \textbf{Raw Feat + Q}: Quantize raw node features into tokens for LLM inference.
    \item + $\mathcal{C}$: Classification using class-specific codebook.
    \end{tablenotes}
    \label{tab:zero_shot}
\end{table}

\subsection{RQ2: Zero-shot Generalization}
\label{sec:zero_shot}
Table~\ref{tab:zero_shot} presents the 5-way zero-shot classification results across different datasets. Unlike existing methods that require labeled source data during pre-training, our framework achieves true zero-shot learning without using labels from either source or target datasets. All models are pre-trained on Cora Full, with OFA requiring labeled data during pre-training. STAG demonstrates strong zero-shot generalization through two inference paths: (1) \emph{Direct LLM inference}: By effectively embedding both semantic and structural information into LLM-compatible formats, STAG achieves significant improvements over baselines, outperforming OFA by large margins on most datasets. This validates the effectiveness of our quantization strategy in bridging continuous graph embeddings and discrete token spaces, enabling STAG to fully leverage the semantic capabilities of LLMs. (2) \emph{Class-specific codebook inference}: As in Eq.~\eqref{eq:c_prediction}, we predict classes by comparing $\boldsymbol{z}_f$ with $\mathcal{C}$. `STAG + $\mathcal{C}$' achieves even stronger performance, demonstrating the quality of our learned representations. The strong performance across both paths, particularly in challenging domain transfers like WikiCS, demonstrates how our framework effectively captures generalizable graph representations. When compared to raw variants, STAG consistently outperforms both 'Raw Feat' with LLM and 'Raw Feat + $\mathcal{C}$' without LLM by effectively incorporating structural information into learned tokens for both inference paths. Complete results on remaining datasets are provided in the Appendix~\ref{sec:appendix_exp}.

\begin{table}[t]
  \centering
  \small
    \caption{5-way 5-shot classification results with different LLMs (pre-trained on Cora Full, $top\text{-}k=5$ for inference).}
  \begin{threeparttable}
  \renewcommand{\arraystretch}{0.9}
  \resizebox{0.95\linewidth}{!}{
  \begin{tabular}{l|cccc}
      \toprule[1.2pt]
      LLM & Cora Full & WikiCS & ogbn-arxiv & CiteSeer \\
      \midrule \midrule
      LLaMA2-7B & 76.66{\scriptsize$\pm$7.79} & 79.00{\scriptsize$\pm$7.96} & 65.33{\scriptsize$\pm$10.46} & 54.35{\scriptsize$\pm$9.54} \\
      + PT & 81.05{\scriptsize$\pm$7.77} & 79.90{\scriptsize$\pm$7.69} & 77.42{\scriptsize$\pm$10.48} & 58.45{\scriptsize$\pm$8.61} \\
      \midrule
      LLaMA2-13B & 77.62{\scriptsize$\pm$8.67} & 79.80{\scriptsize$\pm$7.30} & 69.38{\scriptsize$\pm$8.83} & 54.60{\scriptsize$\pm$8.79} \\
      + PT & 81.95{\scriptsize$\pm$7.06} & 80.45{\scriptsize$\pm$7.66} & 77.75{\scriptsize$\pm$9.01} & 57.30{\scriptsize$\pm$9.20} \\
      \midrule
      Vicuna-7B & 74.12{\scriptsize$\pm$6.47} & 80.30{\scriptsize$\pm$7.02} & 64.84{\scriptsize$\pm$9.38} & 49.25{\scriptsize$\pm$6.72} \\
      + PT & 80.77{\scriptsize$\pm$6.75} & 80.10{\scriptsize$\pm$7.39} & 76.95{\scriptsize$\pm$9.43} & 52.25{\scriptsize$\pm$8.23} \\
      \midrule
      Vicuna-13B & 77.76{\scriptsize$\pm$8.58} & 79.35{\scriptsize$\pm$7.98} & 66.03{\scriptsize$\pm$9.34} & 52.25{\scriptsize$\pm$6.39} \\
      + PT & 81.38{\scriptsize$\pm$7.65} & 79.25{\scriptsize$\pm$7.50} & 75.65{\scriptsize$\pm$9.59} & 53.00{\scriptsize$\pm$8.16} \\
      \midrule
      LLaMA3-8B & 79.22{\scriptsize$\pm$8.45} & 78.40{\scriptsize$\pm$8.05} & 70.37{\scriptsize$\pm$8.95} & 61.25{\scriptsize$\pm$7.14} \\
      + PT & 82.88{\scriptsize$\pm$8.09} & 78.35{\scriptsize$\pm$7.61} & 76.71{\scriptsize$\pm$10.20} & 64.20{\scriptsize$\pm$7.39} \\
      \midrule
      GPT-4o-mini & 79.25{\scriptsize$\pm$8.42} & 81.05{\scriptsize$\pm$6.80} & 71.32{\scriptsize$\pm$9.13} & 61.90{\scriptsize$\pm$7.22} \\
      + PT & 83.04{\scriptsize$\pm$7.84} & \bf 81.90{\scriptsize$\pm$6.16} & 77.51{\scriptsize$\pm$9.58} & 65.90{\scriptsize$\pm$7.04} \\
      \midrule
      GPT-4o & \bf 81.40{\scriptsize$\pm$7.41} & \bf 81.45{\scriptsize$\pm$7.10} & \bf 72.75{\scriptsize$\pm$8.83} & \bf 62.95{\scriptsize$\pm$6.61} \\
      + PT & \bf 83.28{\scriptsize$\pm$7.06} & 81.60{\scriptsize$\pm$7.19} & \bf 78.85{\scriptsize$\pm$9.74} & \bf 65.90{\scriptsize$\pm$7.03} \\
      \bottomrule[1.2pt]
  \end{tabular}}
  \end{threeparttable}
  \begin{tablenotes}
    \footnotesize
    \item $+$ PT: with prompt tuning. 7B, 8B, and 13B indicate the number of parameters in billions.
  \end{tablenotes}
  \label{tab:llm_comparison}
\end{table}

\subsection{RQ3: Flexibility with Different LLMs}
\label{sec:flexibility}

Our quantization strategy enables a unique advantage of STAG: the ability to flexibly pair a single pre-trained model with different LLMs during inference. Unlike existing GraphLLM approaches that require architecture-specific modifications, STAG can seamlessly work with various LLMs, from open-source models like LLaMA2~\cite{touvron2023llama}, LLaMA3~\cite{dubey2024llama}, and Vicuna~\cite{vicuna2023} to closed-source ones like GPT-4o~\cite{hurst2024gpt} by converting graph representations into discrete tokens that are universally interpretable across different LLM architectures. Results in Table~\ref{tab:llm_comparison} demonstrate three key patterns (using $top\text{-}k=5$ for cost-efficient inference with closed-source LLMs): (1) \emph{Model size matters}: Larger models consistently achieve better performance, suggesting enhanced semantic understanding capabilities. (2) \emph{Advanced architectures show advantages}: Newer models like GPT-4o and LLaMA3-8B outperform their predecessors, with STAG's LLM inference becoming competitive even with its traditional variants. (3) \emph{Prompt tuning provides consistent gains}: Performance improves across all LLMs, with particularly notable gains in smaller and older models like LLaMA2-7B and Vicuna-7B, demonstrating the effectiveness and robustness of our adaptation strategy. These results validate our framework's flexibility, offering future-proof advantages as more powerful LLMs become available.

\subsection{Qualitative Analysis}
To provide insights into how STAG processes and represents graph information, we conduct two case studies examining the quantization outputs (using $top\text{-}k=5$ tokens).

\stitle{Structure-Aware Tokenization Analysis.} We analyze a sample node from the `Computational Complexity' class in Cora Full dataset, where our model is pre-trained. This paper, which primarily discusses oracle constructions and isomorphism conjectures, serves as an illustrative example of how STAG effectively integrates structural information into token representations:

\begin{tcolorbox}[boxsep=0mm,left=2.5mm,right=2.5mm]
\footnotesize
\textbf{Raw Text}: ``In this paper we demonstrate an oracle relative to which there are one-way functions but every paddable 1-li-degree collapses to an isomorphism type, thus yielding a relativized failure of the Joseph Young Conjecture [JY85]. We then use this result to construct an oracle relative to which the Isomorphism Conjecture is true but oneway functions exist, which answers an open question of Fenner, Fortnow, and Kurtz [FFK96]. Thus, there are now relativizations realizing every one of the four possible states of affairs between the Isomorphism Conjecture and the existence of one-way functions.''

\textbf{Category}: Computational Complexity

\noindent\rule{\linewidth}{0.4pt}

\textbf{Raw Feat + Quantization}: isomorphism, oracle, numerable, mutable, schemes

\textbf{STAG}: complexity, algebraic, computation, polynomials, compute

\textbf{STAG + PT}: complexity, computational, computation, algorithms, compute
\end{tcolorbox}

This example illustrates the effectiveness of our structure-aware tokenization: While the raw text primarily discusses oracle constructions and isomorphism conjectures, making raw quantization focus on surface-level terms (`isomorphism', `oracle'), STAG successfully captures the paper's theoretical computer science nature by integrating neighborhood information. The learned tokenization emphasizes core complexity theory concepts (`complexity', `computation') and mathematical foundations (`algebraic'), and the prompted version further refines these tokens toward computational complexity aspects. This demonstrates how our method can effectively identify a node's domain even when its raw content is not directly indicative of its category.

\stitle{Intra-class Consistency.} To examine how STAG captures class-specific patterns, we analyze the quantization results for multiple nodes from the same class. The following example presents tokenization outputs for five different nodes from the `Operating Systems' category in WikiCS, showing how our method consistently captures class-relevant features while preserving individual paper characteristics:

\begin{tcolorbox}[boxsep=0mm,left=2.5mm,right=2.5mm]
\footnotesize
\textbf{Node 1}: unix linux, os, kernel, system

\textbf{Node 2}: unix, system, network, interface, protocol

\textbf{Node 3}: unix, system, compiler, terminal, execution

\textbf{Node 4}: unix, linux, kernel, system, filesystem

\textbf{Node 5}: linux, unix, kernel, system, filesystem
\label{box:class_consistency}
\end{tcolorbox}

While all nodes maintain core tokens related to operating systems (`unix', `system'), each node's unique focus is preserved through specific technical tokens (`interface', `terminal'). This demonstrates STAG's ability to balance class-level consistency with instance-level specificity. The consistent appearance of system-related tokens (`kernel', `filesystem') also suggests that our quantization process effectively captures the semantic structure of technical documents beyond simple keyword matching.

\subsection{Ablation Studies}
To validate our design choices in addressing the technical challenges, we conduct ablation studies by systematically removing key components from our framework. All variants are pre-trained on Cora Full. Table~\ref{tab:ablation} presents results on representative datasets, demonstrating each component's contribution: (1) \emph{Semantic and Structural Fusion}: Removing the fusion module (`$\neg$Fusion') to directly quantize node embeddings $\boldsymbol{z}_e$ leads to substantial performance degradation, even when evaluated on the pre-training dataset Cora Full. This underscores the necessity of integrating semantic and structural information to ensure semantic preservation. (2) \emph{KL Regularization}: Eliminating the $\mathcal{L}_{KL}$ (`$\neg\mathcal{L}_{KL}$') results in considerable performance deterioration across both inference paths, confirming its importance in preserving semantic similarity during quantization by encouraging nodes to be mapped to semantically related tokens. (3) \emph{Soft Token Assignment}: Replacing soft assignment with hard assignment (`$\neg$Soft'), which is to quantize into single token during pre-training and inference, severely hampers performance, especially in LLM inference. This demonstrates that soft assignment effectively mitigates overfitting in the quantization process by maintaining distributional information rather than forcing hard decisions, enabling more robust transfer across datasets. These results collectively validate that each component directly addresses a key technical challenge in bridging the gap between graph structures and LLM-compatible representations.

 \begin{table}[t]
  \centering
  \small
    \caption{Ablation study on representative datasets (pre-trained on Cora Full, 5-way 5-shot, $top\text{-}k=13$). 
    }
  \begin{threeparttable}
  \setlength{\tabcolsep}{0.95pt}
  \resizebox{\linewidth}{!}{
  \begin{tabular}{l|ccc|ccc}
      \toprule[1.2pt]
      \multirow{2}{*}{Method} & \multicolumn{3}{c|}{LLM Inference} & \multicolumn{3}{c}{Linear Probing}\\\cline{2-7}
      & Cora Full & WikiCS & ogbn-arxiv & Cora Full & WikiCS & ogbn-arxiv \\
      \midrule \midrule
      Full Model & \bf 80.95{\scriptsize$\pm$8.02} & \bf 79.20{\scriptsize$\pm$8.41} & \bf 71.56{\scriptsize$\pm$10.32} & \bf 86.04{\scriptsize$\pm$6.70} & \bf 84.05{\scriptsize$\pm$5.78} & \bf 82.99{\scriptsize$\pm$8.10} \\
      \midrule
      $\neg$ Fusion & 37.49{\scriptsize$\pm$6.66} & 29.10{\scriptsize$\pm$9.65} & 29.49{\scriptsize$\pm$8.01} & 46.73{\scriptsize$\pm$6.66} & 35.05{\scriptsize$\pm$8.61} & 34.12{\scriptsize$\pm$6.79} \\
      $\neg \mathcal{L}_{KL}$ & 69.74{\scriptsize$\pm$11.04} & 57.95{\scriptsize$\pm$11.48} & 59.79{\scriptsize$\pm$8.32} & 81.83{\scriptsize$\pm$8.33} & 75.05{\scriptsize$\pm$7.21} & 76.18{\scriptsize$\pm$9.21} \\
      $\neg$ Soft & 37.07{\scriptsize$\pm$8.01} & 31.55{\scriptsize$\pm$8.98} & 28.21{\scriptsize$\pm$7.63} & 67.77{\scriptsize$\pm$9.89} & 61.35{\scriptsize$\pm$8.81} & 50.90{\scriptsize$\pm$9.53} \\
      \bottomrule[1.2pt]
  \end{tabular}}
  \end{threeparttable}
  \label{tab:ablation}
\end{table}

\subsection{Computational Efficiency Analysis}
\label{sec:efficiency}

To demonstrate the practical viability of our quantization approach, we analyze the computational overhead of STAG's key components. Our quantization method has time complexity $O(B \times K \times d)$, where $B$ is the total number of nodes in a batch (across all subgraphs), $K$ is the codebook size (15,062 tokens), and $d$ is the embedding dimension (768).

Table~\ref{tab:efficiency} presents a detailed breakdown of computational costs for a typical batch containing 64 subgraphs with approximately 3,944 total nodes. The analysis shows that our quantization process adds minimal overhead compared to the core GNN encoding, representing only 43\% of the total processing time.

\begin{table}[t]
    \centering
    \footnotesize
    \caption{Computational efficiency breakdown of STAG components (pre-trained on Cora Full).}
    \begin{threeparttable}
    \renewcommand{\arraystretch}{1.1}
    \resizebox{0.95\linewidth}{!}{
    \begin{tabular}{l|c|c}
        \toprule[1.2pt]
        Component & Time (ms) & Percentage \\
        \midrule \midrule
        (a) GNN Encoding & 18.4 $\pm$ 41.9 & 57\% \\
        \midrule
        (b) Cosine Similarity (Eq.~\eqref{eq:attn}) & 8.2 $\pm$ 3.1 & 25\% \\
        (c) Weighted Combination (Eq.~\eqref{eq:soft}) & 5.7 $\pm$ 3.3 & 18\% \\
        \midrule
        \textbf{Total Quantization (b+c)} & \textbf{13.9 $\pm$ 6.4} & \textbf{43\%} \\
        \midrule
        \textbf{Total Processing (a+b+c)} & \textbf{32.3 $\pm$ 48.3} & \textbf{100\%} \\
        \bottomrule[1.2pt]
    \end{tabular}}
    \end{threeparttable}
    \label{tab:efficiency}
\end{table}

Compared to traditional GraphLLM approaches that require expensive projector networks for embedding alignment~\cite{tang2024graphgpt}, our frozen codebook design eliminates the need for additional parameter-heavy components during inference. The quantization overhead is dominated by cosine similarity computation with the codebook, which scales linearly with the number of nodes and remains computationally tractable even for large graphs. Our framework achieves a throughput of approximately 121,920 nodes per second, demonstrating that STAG provides a practical solution for real-world deployment scenarios where computational efficiency is crucial.

\subsection{Task Generalization}
\label{sec:task_generalization}

To demonstrate STAG's versatility beyond node classification, we evaluate our framework on two additional graph learning tasks: link prediction and edge classification. All experiments use models pre-trained on source datasets without any task-specific training.

\stitle{Link Prediction.} We evaluate zero-shot binary link prediction following the protocol established by LLaGA~\cite{DChen0JSW24}. Given a pair of nodes, we construct prompts using their quantized tokens and ask the LLM to predict whether an edge exists between them. For the non-LLM variant, we use cosine similarity between node embeddings with a fixed threshold.

Table~\ref{tab:link_prediction} presents the comparison results. Notably, our model achieves comparable or superior performance to LLaGA despite several key differences: (1) STAG is not pre-trained on link prediction tasks, (2) we use only single-dataset pre-training (Arxiv) compared to LLaGA's multi-dataset approach (Arxiv + PubMed + Cora), and (3) our method doesn't use node text during inference, unlike LLaGA.

\begin{table}[t]
    \centering
    \footnotesize
    \caption{Zero-shot binary link prediction comparison with LLaGA.}
    \begin{threeparttable}
    \resizebox{0.75\linewidth}{!}{
    \begin{tabular}{l|c|c}
        \toprule[1.2pt]
        Method & Cora & ogbn-products \\
        \midrule \midrule
        LLaGA & 87.35 & 92.99 \\
        STAG & 63.00 & 92.65 \\
        STAG (non LLM) & \bf 93.20 & \bf 96.85 \\
        \bottomrule[1.2pt]
    \end{tabular}}
    \begin{tablenotes}
    \footnotesize
    \item \centering
    \parbox{0.45\textwidth}{\raggedright
    \textbf{LLaGA} uses multi-dataset pre-training (Arxiv + PubMed for Cora, Arxiv + PubMed + Cora for ogbn-products). \\
    \textbf{STAG} uses single-dataset pre-training (Arxiv only).}
    \end{tablenotes}
    \end{threeparttable}
    \label{tab:link_prediction}
\end{table}

These results highlight STAG's strong generalization capabilities and effective representation learning. The superior performance of our linear probing variant demonstrates that our learned embeddings capture meaningful structural relationships that transfer well to link prediction tasks, even without task-specific training.

\stitle{Edge Classification.} We evaluate edge classification on two text-attributed knowledge graphs: WN18RR (5-way) and FB15K237 (20-way) in N-way 5-shot settings. The LLM is prompted with tokenized (head, tail) pairs to predict the relation type. The non-LLM variant trains a linear classifier on concatenated node embeddings. All models are pre-trained on Cora Full.

\begin{table}[t]
    \centering
    \footnotesize
    \caption{N-way 5-shot edge classification results (pre-trained on Cora Full).}
    \begin{threeparttable}
    \resizebox{0.75\linewidth}{!}{
    \begin{tabular}{l|c|c}
        \toprule[1.2pt]
        Method & WN18RR & FB15K237 \\
        \midrule \midrule
        OFA & 34.35 & 19.55 \\
        STAG & 41.75 & 56.60 \\
        STAG + Linear Probing & \bf 58.30 & \bf 74.80 \\
        \bottomrule[1.2pt]
    \end{tabular}}
    \end{threeparttable}
    \label{tab:edge_classification}
\end{table}

Table~\ref{tab:edge_classification} shows that STAG significantly outperforms OFA across both datasets, demonstrating our framework's effective structural-semantic integration for relation prediction tasks. This superior performance highlights STAG's ability to capture meaningful relational patterns through quantized representations, even when applied to knowledge graph domains that differ substantially from the citation network used for pre-training. The consistent improvements across both WN18RR and FB15K237 validate the generalizability of our quantization approach beyond traditional graph learning tasks.

\section{Conclusion}

We presented STAG, a self-supervised quantization-based framework bridging graph representation learning and LLM. Our framework addresses two fundamental challenges: integrating structural and semantic information while maintaining LLM compatibility, and enabling label-free domain transfer. Through soft token assignment and distribution alignment, STAG effectively integrates structural-semantic information, enables robust cross-dataset transfer, and maintains consistent performance across LLM architectures through a shared vocabulary. Extensive experiments demonstrate STAG's superior performance in both few-shot and zero-shot settings across diverse datasets. Looking forward, STAG could be extended to graph-level tasks like graph classification, as well as link prediction. Additionally, our framework could potentially benefit from advanced LLM architectures and techniques like chain-of-thought reasoning to enhance interpretability and performance. We believe STAG represents a significant step toward more effective and flexible integration of graph learning with LLMs.

\begin{acks}
This research / project is supported by the Ministry of Education, Singapore, under its Academic Research Fund Tier 2 (Proposal ID: T2EP20122-0041). Any opinions, findings and conclusions or recommendations expressed in this material are those of the author(s) and do not reflect the views of the Ministry of Education, Singapore.
\end{acks}

\bibliographystyle{ACM-Reference-Format}
\balance
\bibliography{ref}


\begin{thebibliography}{71}


\ifx \showCODEN    \undefined \def \showCODEN     #1{\unskip}     \fi
\ifx \showISBNx    \undefined \def \showISBNx     #1{\unskip}     \fi
\ifx \showISBNxiii \undefined \def \showISBNxiii  #1{\unskip}     \fi
\ifx \showISSN     \undefined \def \showISSN      #1{\unskip}     \fi
\ifx \showLCCN     \undefined \def \showLCCN      #1{\unskip}     \fi
\ifx \shownote     \undefined \def \shownote      #1{#1}          \fi
\ifx \showarticletitle \undefined \def \showarticletitle #1{#1}   \fi
\ifx \showURL      \undefined \def \showURL       {\relax}        \fi
\providecommand\bibfield[2]{#2}
\providecommand\bibinfo[2]{#2}
\providecommand\natexlab[1]{#1}
\providecommand\showeprint[2][]{arXiv:#2}

\bibitem[Achiam et~al\mbox{.}(2023)]%
        {achiam2023gpt}
\bibfield{author}{\bibinfo{person}{Josh Achiam}, \bibinfo{person}{Steven Adler}, \bibinfo{person}{Sandhini Agarwal}, \bibinfo{person}{Lama Ahmad}, \bibinfo{person}{Ilge Akkaya}, \bibinfo{person}{Florencia~Leoni Aleman}, \bibinfo{person}{Diogo Almeida}, \bibinfo{person}{Janko Altenschmidt}, \bibinfo{person}{Sam Altman}, \bibinfo{person}{Shyamal Anadkat}, {et~al\mbox{.}}} \bibinfo{year}{2023}\natexlab{}.
\newblock \showarticletitle{Gpt-4 technical report}.
\newblock \bibinfo{journal}{\emph{arXiv preprint arXiv:2303.08774}} (\bibinfo{year}{2023}).
\newblock


\bibitem[Akiba et~al\mbox{.}(2019)]%
        {optuna_2019}
\bibfield{author}{\bibinfo{person}{Takuya Akiba}, \bibinfo{person}{Shotaro Sano}, \bibinfo{person}{Toshihiko Yanase}, \bibinfo{person}{Takeru Ohta}, {and} \bibinfo{person}{Masanori Koyama}.} \bibinfo{year}{2019}\natexlab{}.
\newblock \showarticletitle{Optuna: A Next-generation Hyperparameter Optimization Framework}. In \bibinfo{booktitle}{\emph{Proceedings of the 25th {ACM} {SIGKDD} International Conference on Knowledge Discovery and Data Mining}}.
\newblock


\bibitem[Bordes et~al\mbox{.}(2013)]%
        {bordes2013translating}
\bibfield{author}{\bibinfo{person}{Antoine Bordes}, \bibinfo{person}{Nicolas Usunier}, \bibinfo{person}{Alberto Garc{\'{\i}}a{-}Dur{\'{a}}n}, \bibinfo{person}{Jason Weston}, {and} \bibinfo{person}{Oksana Yakhnenko}.} \bibinfo{year}{2013}\natexlab{}.
\newblock \showarticletitle{Translating Embeddings for Modeling Multi-relational Data}. In \bibinfo{booktitle}{\emph{Advances in Neural Information Processing Systems}}, \bibfield{editor}{\bibinfo{person}{Christopher J.~C. Burges}, \bibinfo{person}{L{\'{e}}on Bottou}, \bibinfo{person}{Zoubin Ghahramani}, {and} \bibinfo{person}{Kilian~Q. Weinberger}} (Eds.).
\newblock


\bibitem[Brown et~al\mbox{.}(2020)]%
        {brown2020language}
\bibfield{author}{\bibinfo{person}{Tom Brown}, \bibinfo{person}{Benjamin Mann}, \bibinfo{person}{Nick Ryder}, \bibinfo{person}{Melanie Subbiah}, \bibinfo{person}{Jared~D Kaplan}, \bibinfo{person}{Prafulla Dhariwal}, \bibinfo{person}{Arvind Neelakantan}, \bibinfo{person}{Pranav Shyam}, \bibinfo{person}{Girish Sastry}, \bibinfo{person}{Amanda Askell}, {et~al\mbox{.}}} \bibinfo{year}{2020}\natexlab{}.
\newblock \showarticletitle{Language models are few-shot learners}.
\newblock \bibinfo{journal}{\emph{Advances in neural information processing systems}} (\bibinfo{year}{2020}).
\newblock


\bibitem[Chen et~al\mbox{.}(2024b)]%
        {DChen0JSW24}
\bibfield{author}{\bibinfo{person}{Runjin Chen}, \bibinfo{person}{Tong Zhao}, \bibinfo{person}{Ajay~Kumar Jaiswal}, \bibinfo{person}{Neil Shah}, {and} \bibinfo{person}{Zhangyang Wang}.} \bibinfo{year}{2024}\natexlab{b}.
\newblock \showarticletitle{LLaGA: Large Language and Graph Assistant}. In \bibinfo{booktitle}{\emph{International Conference on Machine Learning,}}.
\newblock


\bibitem[Chen et~al\mbox{.}(2024a)]%
        {chen2024exploring}
\bibfield{author}{\bibinfo{person}{Zhikai Chen}, \bibinfo{person}{Haitao Mao}, \bibinfo{person}{Hang Li}, \bibinfo{person}{Wei Jin}, \bibinfo{person}{Hongzhi Wen}, \bibinfo{person}{Xiaochi Wei}, \bibinfo{person}{Shuaiqiang Wang}, \bibinfo{person}{Dawei Yin}, \bibinfo{person}{Wenqi Fan}, \bibinfo{person}{Hui Liu}, {et~al\mbox{.}}} \bibinfo{year}{2024}\natexlab{a}.
\newblock \showarticletitle{Exploring the potential of large language models (llms) in learning on graphs}.
\newblock \bibinfo{journal}{\emph{ACM SIGKDD Explorations Newsletter}} (\bibinfo{year}{2024}).
\newblock


\bibitem[Chiang et~al\mbox{.}(2023)]%
        {vicuna2023}
\bibfield{author}{\bibinfo{person}{Wei-Lin Chiang}, \bibinfo{person}{Zhuohan Li}, \bibinfo{person}{Zi Lin}, \bibinfo{person}{Ying Sheng}, \bibinfo{person}{Zhanghao Wu}, \bibinfo{person}{Hao Zhang}, \bibinfo{person}{Lianmin Zheng}, \bibinfo{person}{Siyuan Zhuang}, \bibinfo{person}{Yonghao Zhuang}, \bibinfo{person}{Joseph~E. Gonzalez}, \bibinfo{person}{Ion Stoica}, {and} \bibinfo{person}{Eric~P. Xing}.} \bibinfo{year}{2023}\natexlab{}.
\newblock \bibinfo{title}{Vicuna: An Open-Source Chatbot Impressing GPT-4 with 90\%* ChatGPT Quality}.
\newblock
\urldef\tempurl%
\url{https://lmsys.org/blog/2023-03-30-vicuna/}
\showURL{%
\tempurl}


\bibitem[Deng et~al\mbox{.}(2009)]%
        {deng2009imagenet}
\bibfield{author}{\bibinfo{person}{Jia Deng}, \bibinfo{person}{Wei Dong}, \bibinfo{person}{Richard Socher}, \bibinfo{person}{Li-Jia Li}, \bibinfo{person}{Kai Li}, {and} \bibinfo{person}{Li Fei-Fei}.} \bibinfo{year}{2009}\natexlab{}.
\newblock \showarticletitle{Imagenet: A large-scale hierarchical image database}. In \bibinfo{booktitle}{\emph{Proceedings of the IEEE/CVF Conference on Computer Vision and Pattern Recognition}}.
\newblock


\bibitem[Dubey et~al\mbox{.}(2024)]%
        {dubey2024llama}
\bibfield{author}{\bibinfo{person}{Abhimanyu Dubey}, \bibinfo{person}{Abhinav Jauhri}, \bibinfo{person}{Abhinav Pandey}, \bibinfo{person}{Abhishek Kadian}, \bibinfo{person}{Ahmad Al-Dahle}, \bibinfo{person}{Aiesha Letman}, \bibinfo{person}{Akhil Mathur}, \bibinfo{person}{Alan Schelten}, \bibinfo{person}{Amy Yang}, \bibinfo{person}{Angela Fan}, {et~al\mbox{.}}} \bibinfo{year}{2024}\natexlab{}.
\newblock \showarticletitle{The llama 3 herd of models}.
\newblock \bibinfo{journal}{\emph{arXiv preprint arXiv:2407.21783}} (\bibinfo{year}{2024}).
\newblock


\bibitem[Esser et~al\mbox{.}(2021)]%
        {esser2021taming}
\bibfield{author}{\bibinfo{person}{Patrick Esser}, \bibinfo{person}{Robin Rombach}, {and} \bibinfo{person}{Bjorn Ommer}.} \bibinfo{year}{2021}\natexlab{}.
\newblock \showarticletitle{Taming transformers for high-resolution image synthesis}. In \bibinfo{booktitle}{\emph{Proceedings of the IEEE/CVF conference on computer vision and pattern recognition}}.
\newblock


\bibitem[Fan et~al\mbox{.}(2019)]%
        {fan2019graph}
\bibfield{author}{\bibinfo{person}{Wenqi Fan}, \bibinfo{person}{Yao Ma}, \bibinfo{person}{Qing Li}, \bibinfo{person}{Yuan He}, \bibinfo{person}{Eric Zhao}, \bibinfo{person}{Jiliang Tang}, {and} \bibinfo{person}{Dawei Yin}.} \bibinfo{year}{2019}\natexlab{}.
\newblock \showarticletitle{Graph neural networks for social recommendation}. In \bibinfo{booktitle}{\emph{The world wide web conference}}.
\newblock


\bibitem[Fatemi et~al\mbox{.}(2024)]%
        {fatemi2023talk}
\bibfield{author}{\bibinfo{person}{Bahare Fatemi}, \bibinfo{person}{Jonathan Halcrow}, {and} \bibinfo{person}{Bryan Perozzi}.} \bibinfo{year}{2024}\natexlab{}.
\newblock \showarticletitle{Talk like a Graph: Encoding Graphs for Large Language Models}. In \bibinfo{booktitle}{\emph{International Conference on Learning Representations}}.
\newblock


\bibitem[Grover and Leskovec(2016)]%
        {grover2016node2vec}
\bibfield{author}{\bibinfo{person}{Aditya Grover} {and} \bibinfo{person}{Jure Leskovec}.} \bibinfo{year}{2016}\natexlab{}.
\newblock \showarticletitle{node2vec: Scalable feature learning for networks}. In \bibinfo{booktitle}{\emph{Proceedings of the 22nd ACM SIGKDD international conference on Knowledge discovery and data mining}}.
\newblock


\bibitem[Guo et~al\mbox{.}(2025)]%
        {guo2025deepseek}
\bibfield{author}{\bibinfo{person}{Daya Guo}, \bibinfo{person}{Dejian Yang}, \bibinfo{person}{Haowei Zhang}, \bibinfo{person}{Junxiao Song}, \bibinfo{person}{Ruoyu Zhang}, \bibinfo{person}{Runxin Xu}, \bibinfo{person}{Qihao Zhu}, \bibinfo{person}{Shirong Ma}, \bibinfo{person}{Peiyi Wang}, \bibinfo{person}{Xiao Bi}, {et~al\mbox{.}}} \bibinfo{year}{2025}\natexlab{}.
\newblock \showarticletitle{Deepseek-r1: Incentivizing reasoning capability in llms via reinforcement learning}.
\newblock \bibinfo{journal}{\emph{arXiv preprint arXiv:2501.12948}} (\bibinfo{year}{2025}).
\newblock


\bibitem[Guo et~al\mbox{.}(2023)]%
        {guo2023gpt4graph}
\bibfield{author}{\bibinfo{person}{Jiayan Guo}, \bibinfo{person}{Lun Du}, \bibinfo{person}{Hengyu Liu}, \bibinfo{person}{Mengyu Zhou}, \bibinfo{person}{Xinyi He}, {and} \bibinfo{person}{Shi Han}.} \bibinfo{year}{2023}\natexlab{}.
\newblock \showarticletitle{Gpt4graph: Can large language models understand graph structured data? an empirical evaluation and benchmarking}.
\newblock \bibinfo{journal}{\emph{arXiv preprint arXiv:2305.15066}} (\bibinfo{year}{2023}).
\newblock


\bibitem[Hassani and Khasahmadi(2020)]%
        {hassani2020contrastive}
\bibfield{author}{\bibinfo{person}{Kaveh Hassani} {and} \bibinfo{person}{Amir~Hosein Khasahmadi}.} \bibinfo{year}{2020}\natexlab{}.
\newblock \showarticletitle{Contrastive multi-view representation learning on graphs}. In \bibinfo{booktitle}{\emph{International Conference on Machine Learning}}.
\newblock


\bibitem[He et~al\mbox{.}(2021)]%
        {he2020deberta}
\bibfield{author}{\bibinfo{person}{Pengcheng He}, \bibinfo{person}{Xiaodong Liu}, \bibinfo{person}{Jianfeng Gao}, {and} \bibinfo{person}{Weizhu Chen}.} \bibinfo{year}{2021}\natexlab{}.
\newblock \showarticletitle{Deberta: decoding-Enhanced Bert with Disentangled Attention}. In \bibinfo{booktitle}{\emph{International Conference on Learning Representations}}.
\newblock


\bibitem[He et~al\mbox{.}(2024a)]%
        {he2023harnessing}
\bibfield{author}{\bibinfo{person}{Xiaoxin He}, \bibinfo{person}{Xavier Bresson}, \bibinfo{person}{Thomas Laurent}, \bibinfo{person}{Adam Perold}, \bibinfo{person}{Yann LeCun}, {and} \bibinfo{person}{Bryan Hooi}.} \bibinfo{year}{2024}\natexlab{a}.
\newblock \showarticletitle{Harnessing Explanations: LLM-to-LM Interpreter for Enhanced Text-Attributed Graph Representation Learning}. In \bibinfo{booktitle}{\emph{International Conference on Learning Representations}}.
\newblock


\bibitem[He et~al\mbox{.}(2020)]%
        {he2020lightgcn}
\bibfield{author}{\bibinfo{person}{Xiangnan He}, \bibinfo{person}{Kuan Deng}, \bibinfo{person}{Xiang Wang}, \bibinfo{person}{Yan Li}, \bibinfo{person}{Yongdong Zhang}, {and} \bibinfo{person}{Meng Wang}.} \bibinfo{year}{2020}\natexlab{}.
\newblock \showarticletitle{Lightgcn: Simplifying and powering graph convolution network for recommendation}. In \bibinfo{booktitle}{\emph{Proceedings of the 43rd International ACM SIGIR conference on research and development in Information Retrieval}}.
\newblock


\bibitem[He et~al\mbox{.}(2024b)]%
        {he2024g}
\bibfield{author}{\bibinfo{person}{Xiaoxin He}, \bibinfo{person}{Yijun Tian}, \bibinfo{person}{Yifei Sun}, \bibinfo{person}{Nitesh~V. Chawla}, \bibinfo{person}{Thomas Laurent}, \bibinfo{person}{Yann LeCun}, \bibinfo{person}{Xavier Bresson}, {and} \bibinfo{person}{Bryan Hooi}.} \bibinfo{year}{2024}\natexlab{b}.
\newblock \showarticletitle{G-Retriever: Retrieval-Augmented Generation for Textual Graph Understanding and Question Answering}. In \bibinfo{booktitle}{\emph{Advances in Neural Information Processing Systems}}.
\newblock


\bibitem[He et~al\mbox{.}(2024c)]%
        {he2024linkgpt}
\bibfield{author}{\bibinfo{person}{Zhongmou He}, \bibinfo{person}{Jing Zhu}, \bibinfo{person}{Shengyi Qian}, \bibinfo{person}{Joyce Chai}, {and} \bibinfo{person}{Danai Koutra}.} \bibinfo{year}{2024}\natexlab{c}.
\newblock \showarticletitle{LinkGPT: Teaching Large Language Models To Predict Missing Links}.
\newblock \bibinfo{journal}{\emph{arXiv preprint arXiv:2406.04640}} (\bibinfo{year}{2024}).
\newblock


\bibitem[Hoffmann et~al\mbox{.}(2022)]%
        {hoffmann2022training}
\bibfield{author}{\bibinfo{person}{Jordan Hoffmann}, \bibinfo{person}{Sebastian Borgeaud}, \bibinfo{person}{Arthur Mensch}, \bibinfo{person}{Elena Buchatskaya}, \bibinfo{person}{Trevor Cai}, \bibinfo{person}{Eliza Rutherford}, \bibinfo{person}{Diego de~Las Casas}, \bibinfo{person}{Lisa~Anne Hendricks}, \bibinfo{person}{Johannes Welbl}, \bibinfo{person}{Aidan Clark}, {et~al\mbox{.}}} \bibinfo{year}{2022}\natexlab{}.
\newblock \showarticletitle{Training compute-optimal large language models}.
\newblock \bibinfo{journal}{\emph{arXiv preprint arXiv:2203.15556}} (\bibinfo{year}{2022}).
\newblock


\bibitem[Hou et~al\mbox{.}(2023)]%
        {hou2023graphmae2}
\bibfield{author}{\bibinfo{person}{Zhenyu Hou}, \bibinfo{person}{Yufei He}, \bibinfo{person}{Yukuo Cen}, \bibinfo{person}{Xiao Liu}, \bibinfo{person}{Yuxiao Dong}, \bibinfo{person}{Evgeny Kharlamov}, {and} \bibinfo{person}{Jie Tang}.} \bibinfo{year}{2023}\natexlab{}.
\newblock \showarticletitle{Graphmae2: A decoding-enhanced masked self-supervised graph learner}. In \bibinfo{booktitle}{\emph{Proceedings of the ACM web conference}}.
\newblock


\bibitem[Hou et~al\mbox{.}(2022)]%
        {hou2022graphmae}
\bibfield{author}{\bibinfo{person}{Zhenyu Hou}, \bibinfo{person}{Xiao Liu}, \bibinfo{person}{Yukuo Cen}, \bibinfo{person}{Yuxiao Dong}, \bibinfo{person}{Hongxia Yang}, \bibinfo{person}{Chunjie Wang}, {and} \bibinfo{person}{Jie Tang}.} \bibinfo{year}{2022}\natexlab{}.
\newblock \showarticletitle{Graphmae: Self-supervised masked graph autoencoders}. In \bibinfo{booktitle}{\emph{Proceedings of the 28th ACM SIGKDD Conference on Knowledge Discovery and Data Mining}}.
\newblock


\bibitem[Hu et~al\mbox{.}(2020)]%
        {hu2020open}
\bibfield{author}{\bibinfo{person}{Weihua Hu}, \bibinfo{person}{Matthias Fey}, \bibinfo{person}{Marinka Zitnik}, \bibinfo{person}{Yuxiao Dong}, \bibinfo{person}{Hongyu Ren}, \bibinfo{person}{Bowen Liu}, \bibinfo{person}{Michele Catasta}, {and} \bibinfo{person}{Jure Leskovec}.} \bibinfo{year}{2020}\natexlab{}.
\newblock \showarticletitle{Open graph benchmark: Datasets for machine learning on graphs}. In \bibinfo{booktitle}{\emph{Advances in neural information processing systems}}.
\newblock


\bibitem[Huang et~al\mbox{.}(2024)]%
        {huang2024prodigy}
\bibfield{author}{\bibinfo{person}{Qian Huang}, \bibinfo{person}{Hongyu Ren}, \bibinfo{person}{Peng Chen}, \bibinfo{person}{Gregor Kr{\v{z}}manc}, \bibinfo{person}{Daniel Zeng}, \bibinfo{person}{Percy~S Liang}, {and} \bibinfo{person}{Jure Leskovec}.} \bibinfo{year}{2024}\natexlab{}.
\newblock \showarticletitle{Prodigy: Enabling in-context learning over graphs}. In \bibinfo{booktitle}{\emph{Advances in Neural Information Processing Systems}}.
\newblock


\bibitem[Hurst et~al\mbox{.}(2024)]%
        {hurst2024gpt}
\bibfield{author}{\bibinfo{person}{Aaron Hurst}, \bibinfo{person}{Adam Lerer}, \bibinfo{person}{Adam~P Goucher}, \bibinfo{person}{Adam Perelman}, \bibinfo{person}{Aditya Ramesh}, \bibinfo{person}{Aidan Clark}, \bibinfo{person}{AJ Ostrow}, \bibinfo{person}{Akila Welihinda}, \bibinfo{person}{Alan Hayes}, \bibinfo{person}{Alec Radford}, {et~al\mbox{.}}} \bibinfo{year}{2024}\natexlab{}.
\newblock \showarticletitle{Gpt-4o system card}.
\newblock \bibinfo{journal}{\emph{arXiv preprint arXiv:2410.21276}} (\bibinfo{year}{2024}).
\newblock


\bibitem[Jang et~al\mbox{.}(2017)]%
        {jang2016categorical}
\bibfield{author}{\bibinfo{person}{Eric Jang}, \bibinfo{person}{Shixiang Gu}, {and} \bibinfo{person}{Ben Poole}.} \bibinfo{year}{2017}\natexlab{}.
\newblock \showarticletitle{Categorical Reparameterization with Gumbel-Softmax}. In \bibinfo{booktitle}{\emph{International Conference on Learning Representations}}.
\newblock


\bibitem[Kenton and Toutanova(2019)]%
        {kenton2019bert}
\bibfield{author}{\bibinfo{person}{Jacob Devlin Ming-Wei~Chang Kenton} {and} \bibinfo{person}{Lee~Kristina Toutanova}.} \bibinfo{year}{2019}\natexlab{}.
\newblock \showarticletitle{Bert: Pre-training of deep bidirectional transformers for language understanding}. In \bibinfo{booktitle}{\emph{Proceedings of NaacL-HLT}}.
\newblock


\bibitem[Kipf and Welling(2017a)]%
        {kipf2016semi}
\bibfield{author}{\bibinfo{person}{Thomas~N. Kipf} {and} \bibinfo{person}{Max Welling}.} \bibinfo{year}{2017}\natexlab{a}.
\newblock \showarticletitle{Semi-Supervised Classification with Graph Convolutional Networks}. In \bibinfo{booktitle}{\emph{International Conference on Learning Representations}}.
\newblock


\bibitem[Kipf and Welling(2017b)]%
        {thomas2017gcn}
\bibfield{author}{\bibinfo{person}{Thomas~N Kipf} {and} \bibinfo{person}{Max Welling}.} \bibinfo{year}{2017}\natexlab{b}.
\newblock \showarticletitle{Semi-Supervised Classification with Graph Convolutional Networks}. In \bibinfo{booktitle}{\emph{International Conference on Learning Representations}}.
\newblock


\bibitem[Lee et~al\mbox{.}(2022)]%
        {lee2022autoregressive}
\bibfield{author}{\bibinfo{person}{Doyup Lee}, \bibinfo{person}{Chiheon Kim}, \bibinfo{person}{Saehoon Kim}, \bibinfo{person}{Minsu Cho}, {and} \bibinfo{person}{Wook-Shin Han}.} \bibinfo{year}{2022}\natexlab{}.
\newblock \showarticletitle{Autoregressive image generation using residual quantization}. In \bibinfo{booktitle}{\emph{Proceedings of the IEEE/CVF Conference on Computer Vision and Pattern Recognition}}.
\newblock


\bibitem[Li et~al\mbox{.}(2024)]%
        {li2024zerog}
\bibfield{author}{\bibinfo{person}{Yuhan Li}, \bibinfo{person}{Peisong Wang}, \bibinfo{person}{Zhixun Li}, \bibinfo{person}{Jeffrey~Xu Yu}, {and} \bibinfo{person}{Jia Li}.} \bibinfo{year}{2024}\natexlab{}.
\newblock \showarticletitle{Zerog: Investigating cross-dataset zero-shot transferability in graphs}. In \bibinfo{booktitle}{\emph{Proceedings of the 30th ACM SIGKDD Conference on Knowledge Discovery and Data Mining}}.
\newblock


\bibitem[Lin et~al\mbox{.}(2014)]%
        {lin2014microsoft}
\bibfield{author}{\bibinfo{person}{Tsung-Yi Lin}, \bibinfo{person}{Michael Maire}, \bibinfo{person}{Serge Belongie}, \bibinfo{person}{James Hays}, \bibinfo{person}{Pietro Perona}, \bibinfo{person}{Deva Ramanan}, \bibinfo{person}{Piotr Doll{\'a}r}, {and} \bibinfo{person}{C~Lawrence Zitnick}.} \bibinfo{year}{2014}\natexlab{}.
\newblock \showarticletitle{Microsoft coco: Common objects in context}. In \bibinfo{booktitle}{\emph{Computer Vision--ECCV 2014: 13th European Conference}}.
\newblock


\bibitem[Liu et~al\mbox{.}(2024)]%
        {liu2023one}
\bibfield{author}{\bibinfo{person}{Hao Liu}, \bibinfo{person}{Jiarui Feng}, \bibinfo{person}{Lecheng Kong}, \bibinfo{person}{Ningyue Liang}, \bibinfo{person}{Dacheng Tao}, \bibinfo{person}{Yixin Chen}, {and} \bibinfo{person}{Muhan Zhang}.} \bibinfo{year}{2024}\natexlab{}.
\newblock \showarticletitle{One For All: Towards Training One Graph Model For All Classification Tasks}. In \bibinfo{booktitle}{\emph{International Conference on Learning Representations}}.
\newblock


\bibitem[Loshchilov and Hutter(2017)]%
        {Loshchilov2017DecoupledWD}
\bibfield{author}{\bibinfo{person}{Ilya Loshchilov} {and} \bibinfo{person}{Frank Hutter}.} \bibinfo{year}{2017}\natexlab{}.
\newblock \showarticletitle{Decoupled Weight Decay Regularization}. In \bibinfo{booktitle}{\emph{International Conference on Learning Representations}}.
\newblock


\bibitem[Lu et~al\mbox{.}(2021)]%
        {lu2021learning}
\bibfield{author}{\bibinfo{person}{Yuanfu Lu}, \bibinfo{person}{Xunqiang Jiang}, \bibinfo{person}{Yuan Fang}, {and} \bibinfo{person}{Chuan Shi}.} \bibinfo{year}{2021}\natexlab{}.
\newblock \showarticletitle{Learning to pre-train graph neural networks}. In \bibinfo{booktitle}{\emph{Proceedings of the AAAI conference on artificial intelligence}}.
\newblock


\bibitem[McAuley et~al\mbox{.}(2015)]%
        {mcauley2015image}
\bibfield{author}{\bibinfo{person}{Julian McAuley}, \bibinfo{person}{Christopher Targett}, \bibinfo{person}{Qinfeng Shi}, {and} \bibinfo{person}{Anton Van Den~Hengel}.} \bibinfo{year}{2015}\natexlab{}.
\newblock \showarticletitle{Image-based recommendations on styles and substitutes}. In \bibinfo{booktitle}{\emph{Proceedings of the 38th international ACM SIGIR conference on research and development in information retrieval}}.
\newblock


\bibitem[McCallum et~al\mbox{.}(2000)]%
        {mccallum2000automating}
\bibfield{author}{\bibinfo{person}{Andrew~Kachites McCallum}, \bibinfo{person}{Kamal Nigam}, \bibinfo{person}{Jason Rennie}, {and} \bibinfo{person}{Kristie Seymore}.} \bibinfo{year}{2000}\natexlab{}.
\newblock \showarticletitle{Automating the construction of internet portals with machine learning}.
\newblock \bibinfo{journal}{\emph{Information Retrieval}} (\bibinfo{year}{2000}).
\newblock


\bibitem[Mernyei and Cangea(2020)]%
        {mernyei2020wiki}
\bibfield{author}{\bibinfo{person}{P{\'e}ter Mernyei} {and} \bibinfo{person}{C{\u{a}}t{\u{a}}lina Cangea}.} \bibinfo{year}{2020}\natexlab{}.
\newblock \showarticletitle{Wiki-CS: A Wikipedia-Based Benchmark for Graph Neural Networks}.
\newblock \bibinfo{journal}{\emph{arXiv preprint arXiv:2007.02901}} (\bibinfo{year}{2020}).
\newblock


\bibitem[Oord et~al\mbox{.}(2018)]%
        {oord2018representation}
\bibfield{author}{\bibinfo{person}{Aaron van~den Oord}, \bibinfo{person}{Yazhe Li}, {and} \bibinfo{person}{Oriol Vinyals}.} \bibinfo{year}{2018}\natexlab{}.
\newblock \showarticletitle{Representation learning with contrastive predictive coding}.
\newblock \bibinfo{journal}{\emph{arXiv preprint arXiv:1807.03748}} (\bibinfo{year}{2018}).
\newblock


\bibitem[Perozzi et~al\mbox{.}(2014)]%
        {perozzi2014deepwalk}
\bibfield{author}{\bibinfo{person}{Bryan Perozzi}, \bibinfo{person}{Rami Al-Rfou}, {and} \bibinfo{person}{Steven Skiena}.} \bibinfo{year}{2014}\natexlab{}.
\newblock \showarticletitle{Deepwalk: Online learning of social representations}. In \bibinfo{booktitle}{\emph{Proceedings of the 20th ACM SIGKDD international conference on Knowledge discovery and data mining}}.
\newblock


\bibitem[Qu et~al\mbox{.}(2024)]%
        {qu2024llms}
\bibfield{author}{\bibinfo{person}{Haoxuan Qu}, \bibinfo{person}{Yujun Cai}, {and} \bibinfo{person}{Jun Liu}.} \bibinfo{year}{2024}\natexlab{}.
\newblock \showarticletitle{Llms are good action recognizers}. In \bibinfo{booktitle}{\emph{Proceedings of the IEEE/CVF Conference on Computer Vision and Pattern Recognition}}.
\newblock


\bibitem[Ramesh et~al\mbox{.}(2021)]%
        {ramesh2021zero}
\bibfield{author}{\bibinfo{person}{Aditya Ramesh}, \bibinfo{person}{Mikhail Pavlov}, \bibinfo{person}{Gabriel Goh}, \bibinfo{person}{Scott Gray}, \bibinfo{person}{Chelsea Voss}, \bibinfo{person}{Alec Radford}, \bibinfo{person}{Mark Chen}, {and} \bibinfo{person}{Ilya Sutskever}.} \bibinfo{year}{2021}\natexlab{}.
\newblock \showarticletitle{Zero-shot text-to-image generation}. In \bibinfo{booktitle}{\emph{International conference on machine learning}}.
\newblock


\bibitem[Razavi et~al\mbox{.}(2019)]%
        {razavi2019generating}
\bibfield{author}{\bibinfo{person}{Ali Razavi}, \bibinfo{person}{Aaron Van~den Oord}, {and} \bibinfo{person}{Oriol Vinyals}.} \bibinfo{year}{2019}\natexlab{}.
\newblock \showarticletitle{Generating diverse high-fidelity images with vq-vae-2}. In \bibinfo{booktitle}{\emph{Advances in neural information processing systems}}.
\newblock


\bibitem[Reimers and Gurevych(2019)]%
        {Reimers2019SentenceBERTSE}
\bibfield{author}{\bibinfo{person}{Nils Reimers} {and} \bibinfo{person}{Iryna Gurevych}.} \bibinfo{year}{2019}\natexlab{}.
\newblock \showarticletitle{Sentence-BERT: Sentence Embeddings using Siamese BERT-Networks}. In \bibinfo{booktitle}{\emph{Conference on Empirical Methods in Natural Language Processing}}.
\newblock


\bibitem[Rombach et~al\mbox{.}(2022)]%
        {rombach2022high}
\bibfield{author}{\bibinfo{person}{Robin Rombach}, \bibinfo{person}{Andreas Blattmann}, \bibinfo{person}{Dominik Lorenz}, \bibinfo{person}{Patrick Esser}, {and} \bibinfo{person}{Bj{\"o}rn Ommer}.} \bibinfo{year}{2022}\natexlab{}.
\newblock \showarticletitle{High-resolution image synthesis with latent diffusion models}. In \bibinfo{booktitle}{\emph{Proceedings of the IEEE/CVF conference on computer vision and pattern recognition}}.
\newblock


\bibitem[Sun et~al\mbox{.}(2022)]%
        {sun2022gppt}
\bibfield{author}{\bibinfo{person}{Mingchen Sun}, \bibinfo{person}{Kaixiong Zhou}, \bibinfo{person}{Xin He}, \bibinfo{person}{Ying Wang}, {and} \bibinfo{person}{Xin Wang}.} \bibinfo{year}{2022}\natexlab{}.
\newblock \showarticletitle{Gppt: Graph pre-training and prompt tuning to generalize graph neural networks}. In \bibinfo{booktitle}{\emph{Proceedings of the 28th ACM SIGKDD Conference on Knowledge Discovery and Data Mining}}.
\newblock


\bibitem[Sun et~al\mbox{.}(2023)]%
        {sun2023all}
\bibfield{author}{\bibinfo{person}{Xiangguo Sun}, \bibinfo{person}{Hong Cheng}, \bibinfo{person}{Jia Li}, \bibinfo{person}{Bo Liu}, {and} \bibinfo{person}{Jihong Guan}.} \bibinfo{year}{2023}\natexlab{}.
\newblock \showarticletitle{All in one: Multi-task prompting for graph neural networks}. In \bibinfo{booktitle}{\emph{Proceedings of the 29th ACM SIGKDD Conference on Knowledge Discovery and Data Mining}}.
\newblock


\bibitem[Tang et~al\mbox{.}(2024)]%
        {tang2024graphgpt}
\bibfield{author}{\bibinfo{person}{Jiabin Tang}, \bibinfo{person}{Yuhao Yang}, \bibinfo{person}{Wei Wei}, \bibinfo{person}{Lei Shi}, \bibinfo{person}{Lixin Su}, \bibinfo{person}{Suqi Cheng}, \bibinfo{person}{Dawei Yin}, {and} \bibinfo{person}{Chao Huang}.} \bibinfo{year}{2024}\natexlab{}.
\newblock \showarticletitle{Graphgpt: Graph instruction tuning for large language models}. In \bibinfo{booktitle}{\emph{Proceedings of the 47th International ACM SIGIR Conference on Research and Development in Information Retrieval}}.
\newblock


\bibitem[Thakoor et~al\mbox{.}(2022)]%
        {thakoor2022large}
\bibfield{author}{\bibinfo{person}{Shantanu Thakoor}, \bibinfo{person}{Corentin Tallec}, \bibinfo{person}{Mohammad~Gheshlaghi Azar}, \bibinfo{person}{Mehdi Azabou}, \bibinfo{person}{Eva~L Dyer}, \bibinfo{person}{Remi Munos}, \bibinfo{person}{Petar Veli{\v{c}}kovi{\'c}}, {and} \bibinfo{person}{Michal Valko}.} \bibinfo{year}{2022}\natexlab{}.
\newblock \showarticletitle{Large-scale representation learning on graphs via bootstrapping}. In \bibinfo{booktitle}{\emph{International Conference on Learning Representations}}.
\newblock


\bibitem[Tian et~al\mbox{.}(2024)]%
        {tian2024graph}
\bibfield{author}{\bibinfo{person}{Yijun Tian}, \bibinfo{person}{Huan Song}, \bibinfo{person}{Zichen Wang}, \bibinfo{person}{Haozhu Wang}, \bibinfo{person}{Ziqing Hu}, \bibinfo{person}{Fang Wang}, \bibinfo{person}{Nitesh~V Chawla}, {and} \bibinfo{person}{Panpan Xu}.} \bibinfo{year}{2024}\natexlab{}.
\newblock \showarticletitle{Graph neural prompting with large language models}. In \bibinfo{booktitle}{\emph{Proceedings of the AAAI Conference on Artificial Intelligence}}.
\newblock


\bibitem[Touvron et~al\mbox{.}(2023)]%
        {touvron2023llama}
\bibfield{author}{\bibinfo{person}{Hugo Touvron}, \bibinfo{person}{Louis Martin}, \bibinfo{person}{Kevin Stone}, \bibinfo{person}{Peter Albert}, \bibinfo{person}{Amjad Almahairi}, \bibinfo{person}{Yasmine Babaei}, \bibinfo{person}{Nikolay Bashlykov}, \bibinfo{person}{Soumya Batra}, \bibinfo{person}{Prajjwal Bhargava}, \bibinfo{person}{Shruti Bhosale}, {et~al\mbox{.}}} \bibinfo{year}{2023}\natexlab{}.
\newblock \showarticletitle{Llama 2: Open foundation and fine-tuned chat models}.
\newblock \bibinfo{journal}{\emph{arXiv preprint arXiv:2307.09288}} (\bibinfo{year}{2023}).
\newblock


\bibitem[Trouillon et~al\mbox{.}(2016)]%
        {trouillon2016complex}
\bibfield{author}{\bibinfo{person}{Th{\'e}o Trouillon}, \bibinfo{person}{Johannes Welbl}, \bibinfo{person}{Sebastian Riedel}, \bibinfo{person}{{\'E}ric Gaussier}, {and} \bibinfo{person}{Guillaume Bouchard}.} \bibinfo{year}{2016}\natexlab{}.
\newblock \showarticletitle{Complex embeddings for simple link prediction}. In \bibinfo{booktitle}{\emph{International conference on machine learning}}.
\newblock


\bibitem[van~den Oord et~al\mbox{.}(2017)]%
        {van2017neural}
\bibfield{author}{\bibinfo{person}{A{\"{a}}ron van~den Oord}, \bibinfo{person}{Oriol Vinyals}, {and} \bibinfo{person}{Koray Kavukcuoglu}.} \bibinfo{year}{2017}\natexlab{}.
\newblock \showarticletitle{Neural Discrete Representation Learning}. In \bibinfo{booktitle}{\emph{Advances in Neural Information Processing Systems}}.
\newblock


\bibitem[Veli{\v{c}}kovi{\'c} et~al\mbox{.}(2018)]%
        {velivckovic2018graph}
\bibfield{author}{\bibinfo{person}{Petar Veli{\v{c}}kovi{\'c}}, \bibinfo{person}{Guillem Cucurull}, \bibinfo{person}{Arantxa Casanova}, \bibinfo{person}{Adriana Romero}, \bibinfo{person}{Pietro Li{\`o}}, {and} \bibinfo{person}{Yoshua Bengio}.} \bibinfo{year}{2018}\natexlab{}.
\newblock \showarticletitle{Graph Attention Networks}. In \bibinfo{booktitle}{\emph{International Conference on Learning Representations}}.
\newblock


\bibitem[Veličković et~al\mbox{.}(2019)]%
        {velickovic2019deep}
\bibfield{author}{\bibinfo{person}{Petar Veličković}, \bibinfo{person}{William Fedus}, \bibinfo{person}{William~L. Hamilton}, \bibinfo{person}{Pietro Liò}, \bibinfo{person}{Yoshua Bengio}, {and} \bibinfo{person}{R.~Devon Hjelm}.} \bibinfo{year}{2019}\natexlab{}.
\newblock \showarticletitle{Deep Graph Infomax}. In \bibinfo{booktitle}{\emph{International Conference on Learning Representations}}.
\newblock


\bibitem[Wang et~al\mbox{.}(2024)]%
        {wang2024can}
\bibfield{author}{\bibinfo{person}{Heng Wang}, \bibinfo{person}{Shangbin Feng}, \bibinfo{person}{Tianxing He}, \bibinfo{person}{Zhaoxuan Tan}, \bibinfo{person}{Xiaochuang Han}, {and} \bibinfo{person}{Yulia Tsvetkov}.} \bibinfo{year}{2024}\natexlab{}.
\newblock \showarticletitle{Can language models solve graph problems in natural language?}. In \bibinfo{booktitle}{\emph{Advances in Neural Information Processing Systems}}.
\newblock


\bibitem[Wang et~al\mbox{.}(2019)]%
        {wang2019neural}
\bibfield{author}{\bibinfo{person}{Xiang Wang}, \bibinfo{person}{Xiangnan He}, \bibinfo{person}{Meng Wang}, \bibinfo{person}{Fuli Feng}, {and} \bibinfo{person}{Tat-Seng Chua}.} \bibinfo{year}{2019}\natexlab{}.
\newblock \showarticletitle{Neural graph collaborative filtering}. In \bibinfo{booktitle}{\emph{Proceedings of the 42nd international ACM SIGIR conference on Research and development in Information Retrieval}}.
\newblock


\bibitem[Wang et~al\mbox{.}(2014)]%
        {wang2014knowledge}
\bibfield{author}{\bibinfo{person}{Zhen Wang}, \bibinfo{person}{Jianwen Zhang}, \bibinfo{person}{Jianlin Feng}, {and} \bibinfo{person}{Zheng Chen}.} \bibinfo{year}{2014}\natexlab{}.
\newblock \showarticletitle{Knowledge graph embedding by translating on hyperplanes}. In \bibinfo{booktitle}{\emph{Proceedings of the AAAI conference on artificial intelligence}}.
\newblock


\bibitem[Wen and Fang(2024)]%
        {wen2024prompt}
\bibfield{author}{\bibinfo{person}{Zhihao Wen} {and} \bibinfo{person}{Yuan Fang}.} \bibinfo{year}{2024}\natexlab{}.
\newblock \showarticletitle{Prompt tuning on graph-augmented low-resource text classification}.
\newblock \bibinfo{journal}{\emph{IEEE Transactions on Knowledge and Data Engineering}} (\bibinfo{year}{2024}).
\newblock


\bibitem[Xie et~al\mbox{.}(2022)]%
        {xie2022self}
\bibfield{author}{\bibinfo{person}{Yaochen Xie}, \bibinfo{person}{Zhao Xu}, {and} \bibinfo{person}{Shuiwang Ji}.} \bibinfo{year}{2022}\natexlab{}.
\newblock \showarticletitle{Self-supervised representation learning via latent graph prediction}. In \bibinfo{booktitle}{\emph{International Conference on Machine Learning}}.
\newblock


\bibitem[Yan et~al\mbox{.}(2023)]%
        {yan2023comprehensive}
\bibfield{author}{\bibinfo{person}{Hao Yan}, \bibinfo{person}{Chaozhuo Li}, \bibinfo{person}{Ruosong Long}, \bibinfo{person}{Chao Yan}, \bibinfo{person}{Jianan Zhao}, \bibinfo{person}{Wenwen Zhuang}, \bibinfo{person}{Jun Yin}, \bibinfo{person}{Peiyan Zhang}, \bibinfo{person}{Weihao Han}, \bibinfo{person}{Hao Sun}, \bibinfo{person}{Weiwei Deng}, \bibinfo{person}{Qi Zhang}, \bibinfo{person}{Lichao Sun}, \bibinfo{person}{Xing Xie}, {and} \bibinfo{person}{Senzhang Wang}.} \bibinfo{year}{2023}\natexlab{}.
\newblock \showarticletitle{A Comprehensive Study on Text-attributed Graphs: Benchmarking and Rethinking}. In \bibinfo{booktitle}{\emph{Advances in Neural Information Processing Systems}}.
\newblock


\bibitem[Yang et~al\mbox{.}(2024)]%
        {yang2024qwen2}
\bibfield{author}{\bibinfo{person}{An Yang}, \bibinfo{person}{Baosong Yang}, \bibinfo{person}{Beichen Zhang}, \bibinfo{person}{Binyuan Hui}, \bibinfo{person}{Bo Zheng}, \bibinfo{person}{Bowen Yu}, \bibinfo{person}{Chengyuan Li}, \bibinfo{person}{Dayiheng Liu}, \bibinfo{person}{Fei Huang}, \bibinfo{person}{Haoran Wei}, {et~al\mbox{.}}} \bibinfo{year}{2024}\natexlab{}.
\newblock \showarticletitle{Qwen2. 5 technical report}.
\newblock \bibinfo{journal}{\emph{arXiv preprint arXiv:2412.15115}} (\bibinfo{year}{2024}).
\newblock


\bibitem[Yang et~al\mbox{.}(2016)]%
        {yang2016revisiting}
\bibfield{author}{\bibinfo{person}{Zhilin Yang}, \bibinfo{person}{William Cohen}, {and} \bibinfo{person}{Ruslan Salakhudinov}.} \bibinfo{year}{2016}\natexlab{}.
\newblock \showarticletitle{Revisiting semi-supervised learning with graph embeddings}. In \bibinfo{booktitle}{\emph{Proceedings of the 33rd International Conference on Machine Learning (ICML)}}.
\newblock


\bibitem[Ye et~al\mbox{.}(2023)]%
        {ye2023natural}
\bibfield{author}{\bibinfo{person}{Ruosong Ye}, \bibinfo{person}{Caiqi Zhang}, \bibinfo{person}{Runhui Wang}, \bibinfo{person}{Shuyuan Xu}, \bibinfo{person}{Yongfeng Zhang}, {et~al\mbox{.}}} \bibinfo{year}{2023}\natexlab{}.
\newblock \showarticletitle{Natural language is all a graph needs}.
\newblock \bibinfo{journal}{\emph{arXiv preprint arXiv:2308.07134}} (\bibinfo{year}{2023}).
\newblock


\bibitem[Zeghidour et~al\mbox{.}(2021)]%
        {zeghidour2021soundstream}
\bibfield{author}{\bibinfo{person}{Neil Zeghidour}, \bibinfo{person}{Alejandro Luebs}, \bibinfo{person}{Ahmed Omran}, \bibinfo{person}{Jan Skoglund}, {and} \bibinfo{person}{Marco Tagliasacchi}.} \bibinfo{year}{2021}\natexlab{}.
\newblock \showarticletitle{Soundstream: An end-to-end neural audio codec}.
\newblock \bibinfo{journal}{\emph{IEEE/ACM Transactions on Audio, Speech, and Language Processing}} (\bibinfo{year}{2021}).
\newblock


\bibitem[Zhang et~al\mbox{.}(2021)]%
        {zhang2021canonical}
\bibfield{author}{\bibinfo{person}{Hengrui Zhang}, \bibinfo{person}{Qitian Wu}, \bibinfo{person}{Junchi Yan}, \bibinfo{person}{David Wipf}, {and} \bibinfo{person}{Philip~S Yu}.} \bibinfo{year}{2021}\natexlab{}.
\newblock \showarticletitle{From canonical correlation analysis to self-supervised graph neural networks}. In \bibinfo{booktitle}{\emph{Advances in Neural Information Processing Systems}}.
\newblock


\bibitem[Zhao et~al\mbox{.}(2023a)]%
        {zhao2023gimlet}
\bibfield{author}{\bibinfo{person}{Haiteng Zhao}, \bibinfo{person}{Shengchao Liu}, \bibinfo{person}{Chang Ma}, \bibinfo{person}{Hannan Xu}, \bibinfo{person}{Jie Fu}, \bibinfo{person}{Zhihong Deng}, \bibinfo{person}{Lingpeng Kong}, {and} \bibinfo{person}{Qi Liu}.} \bibinfo{year}{2023}\natexlab{a}.
\newblock \showarticletitle{{GIMLET:} {A} Unified Graph-Text Model for Instruction-Based Molecule Zero-Shot Learning}. In \bibinfo{booktitle}{\emph{Advances in Neural Information Processing Systems}}.
\newblock


\bibitem[Zhao et~al\mbox{.}(2023b)]%
        {zhao2023graphtext}
\bibfield{author}{\bibinfo{person}{Jianan Zhao}, \bibinfo{person}{Le Zhuo}, \bibinfo{person}{Yikang Shen}, \bibinfo{person}{Meng Qu}, \bibinfo{person}{Kai Liu}, \bibinfo{person}{Michael Bronstein}, \bibinfo{person}{Zhaocheng Zhu}, {and} \bibinfo{person}{Jian Tang}.} \bibinfo{year}{2023}\natexlab{b}.
\newblock \showarticletitle{Graphtext: Graph reasoning in text space}.
\newblock \bibinfo{journal}{\emph{arXiv preprint arXiv:2310.01089}} (\bibinfo{year}{2023}).
\newblock


\bibitem[Zhu et~al\mbox{.}(2024)]%
        {zhu2024beyond}
\bibfield{author}{\bibinfo{person}{Lei Zhu}, \bibinfo{person}{Fangyun Wei}, {and} \bibinfo{person}{Yanye Lu}.} \bibinfo{year}{2024}\natexlab{}.
\newblock \showarticletitle{Beyond text: Frozen large language models in visual signal comprehension}. In \bibinfo{booktitle}{\emph{Proceedings of the IEEE/CVF Conference on Computer Vision and Pattern Recognition}}.
\newblock


\end{thebibliography}

\appendix
\label{sec:appendix}

\begin{table}[H]
    \centering
    \footnotesize
    \caption{Hyperparameter configurations for model pre-training on different datasets.}
    \begin{threeparttable}
    \renewcommand{\arraystretch}{0.9}
    \resizebox{0.95\linewidth}{!}{
    \begin{tabular}{l|ccc}
    \toprule[1.2pt]
     Dataset & Cora Full & ogbn-arxiv & ogbn-products\\
     \midrule \midrule
     mask rate & 0.53 & 0.6 & 0.74 \\
     learning rate & $5.0\times 10^{-5}$ & $2.32\times 10^{-4}$ & $3.47\times 10^{-4}$\\
     weight decay  & $1.88\times 10^{-6}$ & $9.94\times 10^{-3}$ & $1.57\times 10^{-3}$ \\
     epoch & 20 & 16 & 10 \\
    activation & elu & prelu & relu \\
     hidden dim & 256 & 512 & 512 \\
     \# layers & 3 & 1 & 2 \\
     \# heads & 2 & 2 & 4 \\
    \# neg & 20 & 23 & 16 \\
     $\tau_c$ & 0.831 & 0.354 & 0.103 \\
     $\beta$ & 1.9 & 0.58 & 1.4 \\
     $\lambda$ & 1.0 & 1.0 & 1.6 \\
    \bottomrule[1.2pt]
    \end{tabular}}
    \end{threeparttable}
    \label{tab:pretrain_configs}
\end{table}

\begin{table*}[t]
  \centering
  \small
    \caption{5-way 5-shot node classification across different datasets (except PubMed: 3-way). Models are pre-trained on ogbn-arxiv with $top\text{-}k=13$ for LLM inference. Colored cells: pre-train dataset matches target (red: ogbn-arxiv). Results show accuracy (\%) averaged over 20 random tasks, with best results among our variants in bold.}
  \begin{threeparttable}
  \resizebox{0.99\linewidth}{!}{
  \begin{tabular}{l|c|c|ccccccc}
      \toprule[1.2pt]
      \multirow{2}{*}{Pre-train data} & \multirow{2}{*}{Method} & \multirow{2}{*}{LLM} & \multicolumn{7}{c}{Target data}\\\cline{4-10}
       &   &   & Cora & Cora Full & CiteSeer & PubMed & WikiCS & ogbn-arxiv & ogbn-products \\
      \midrule\midrule
      \multirow{11}{*}{ogbn-arxiv}
      & DGI & \ding{55} & OOM & OOM & OOM & OOM & OOM & \cellcolor{red!20}OOM & OOM \\
      & GraphMAE2 & \ding{55} & 78.85{\scriptsize$\pm$4.78} & 83.54{\scriptsize$\pm$7.63} & 63.35{\scriptsize$\pm$5.82} & 65.90{\scriptsize$\pm$6.82} & 76.90{\scriptsize$\pm$7.70} & \cellcolor{red!20}79.61{\scriptsize$\pm$8.04} & 74.11{\scriptsize$\pm$8.73} \\
        \noalign{\vskip -2pt}  
        \cmidrule{2-10}
        \noalign{\vskip -2pt}  
      & GPPT & \ding{55} & 25.80{\scriptsize$\pm$5.20} & 32.95{\scriptsize$\pm$10.99} & 28.80{\scriptsize$\pm$6.20} & 18.28{\scriptsize$\pm$11.14} & 25.05{\scriptsize$\pm$7.84} & \cellcolor{red!20}64.30{\scriptsize$\pm$13.19} & 22.25{\scriptsize$\pm$3.99} \\
      & G2P2 & \ding{55} & 74.10{\scriptsize$\pm$7.00} & 80.50{\scriptsize$\pm$7.07} & 58.90{\scriptsize$\pm$9.66} & 66.20{\scriptsize$\pm$8.02} & 70.25{\scriptsize$\pm$8.14} & \cellcolor{red!20}71.30{\scriptsize$\pm$9.15} & 71.54{\scriptsize$\pm$10.75} \\
        \noalign{\vskip -2pt}  
        \cmidrule{2-10}
        \noalign{\vskip -2pt}  
      & Prodigy & \ding{55} & 48.60{\scriptsize$\pm$6.12} & 60.90{\scriptsize$\pm$7.89} & 44.45{\scriptsize$\pm$6.30} & 55.27{\scriptsize$\pm$7.22} & 57.40{\scriptsize$\pm$6.33} & \cellcolor{red!20}46.10{\scriptsize$\pm$5.67} & 41.50{\scriptsize$\pm$6.85} \\
      & OFA & \ding{55} & 49.90{\scriptsize$\pm$5.67} & 64.85{\scriptsize$\pm$3.82} & 51.40{\scriptsize$\pm$6.05} & 42.40{\scriptsize$\pm$3.68} & 51.90{\scriptsize$\pm$6.15} & \cellcolor{red!20}65.00{\scriptsize$\pm$3.54} & 42.50{\scriptsize$\pm$5.01} \\
        \noalign{\vskip -2pt}  
        \cmidrule{2-10}
        \noalign{\vskip -2pt}  
      & STAG & \ding{51} & 63.50{\scriptsize$\pm$7.85} & 79.32{\scriptsize$\pm$7.47} & 60.05{\scriptsize$\pm$6.61} & 55.00{\scriptsize$\pm$7.06} & 79.10{\scriptsize$\pm$8.37} & \cellcolor{red!20}71.99{\scriptsize$\pm$7.89} & 70.05{\scriptsize$\pm$9.30} \\
    & + Linear Probing & \ding{55} & \bf 76.95{\scriptsize$\pm$5.91} & 84.83{\scriptsize$\pm$7.22} & 65.05{\scriptsize$\pm$6.26} & 69.70{\scriptsize$\pm$6.73} & 82.05{\scriptsize$\pm$6.96} & \cellcolor{red!20}82.71{\scriptsize$\pm$8.11} & 78.22{\scriptsize$\pm$8.14} \\
      & + Prompt Tuning & \ding{51} & 71.45{\scriptsize$\pm$4.97} & 83.05{\scriptsize$\pm$7.95} & 62.65{\scriptsize$\pm$5.07} & 62.90{\scriptsize$\pm$5.86} & 81.55{\scriptsize$\pm$7.55} & \cellcolor{red!20}79.28{\scriptsize$\pm$8.62} & 71.93{\scriptsize$\pm$7.02} \\
      & + Prompt Tuning* & \ding{55} & 76.75{\scriptsize$\pm$5.78} & \bf 85.82{\scriptsize$\pm$8.01} & \bf 65.40{\scriptsize$\pm$6.27} & \bf 70.05{\scriptsize$\pm$5.49} & \bf 83.15{\scriptsize$\pm$7.09} & \cellcolor{red!20}\bf 82.91{\scriptsize$\pm$7.69} & \bf 79.40{\scriptsize$\pm$7.43} \\
      \bottomrule[1.2pt]
  \end{tabular}}
  \begin{tablenotes}
      \footnotesize
      \item \ding{51}/\ding{55}: LLM usage during inference; \textbf{Prompt Tuning*}: inference without LLM; \textbf{Raw Text}: Use raw text for LLM inference (To fit in the context window of LLM, raw text is truncated); \textbf{Raw Feat + Quantization}: Directly quantize raw node features into tokens for LLM inference; \textbf{Raw Feat + Linear Probing}: Train a linear classifier on raw node features without any pre-training; \textbf{OOM}: Out-of-memory error during training.
  \end{tablenotes}
  \end{threeparttable}
  \label{tab:few_shot_arxiv}
\end{table*}

\section*{Appendices}
\section{Implementation Details}
\label{sec:appendix_implementation}
\subsection{Hardware Configuration}
In our experiments, we used a Linux machine equipped with an AMD EPYC 7763 64-core processor (3.53 GHz) and an NVIDIA L40 GPU (40 GB).

\subsection{Model Configuration}
Our model is pre-trained using the Adamw optimizer~\cite{Loshchilov2017DecoupledWD}, and we employ the Tree-structured Parzen Estimator (TPE) from Optuna~\cite{optuna_2019} for hyperparameter optimization. The codebook $\boldsymbol{E}$ has 15,062 tokens across all datasets. The detailed hyperparameter settings for model pretraining are provided in Table~\ref{tab:pretrain_configs}.

\section{Supplementary Experiments and Analysis}
\label{sec:appendix_exp}

\subsection{Few-shot Learning Performance}
\label{sec:appendix_few_shot}
Table~\ref{tab:few_shot_arxiv} presents the few-shot learning results when pre-trained on ogbn-arxiv. And STAG performance trends remain consistent with Cora Full pre-training results.

\begin{table}[t]
    \centering
    \small
    \caption{5-way zero-shot node classification results (except PubMed with 3 classes). We report accuracy (\%) averaged over 20 random tasks with standard deviation.}
    \begin{threeparttable}
    \setlength{\tabcolsep}{0.9pt}
    \resizebox{\linewidth}{!}{
    \begin{tabular}{l|c|c|ccc}
        \toprule[1.2pt]
        \multirow{2}{*}{Pre-train data} & \multirow{2}{*}{Method} & \multirow{2}{*}{LLM} & \multicolumn{3}{c}{Target data}\\\cline{4-6}
        &   &   & CiteSeer & PubMed  & ogbn-products \\
        \midrule \midrule
        \multirow{2}{*}{No pre-train}& Raw Feat + Q & \ding{51} & 43.90{\scriptsize$\pm$5.55} & 42.35{\scriptsize$\pm$2.95} & 62.42{\scriptsize$\pm$10.84} \\
        & Raw Feat + $\mathcal{C}$ & \ding{55} & 62.35{\scriptsize$\pm$6.55} & 63.50{\scriptsize$\pm$4.14} & 74.66{\scriptsize$\pm$7.68} \\
        \midrule
        \multirow{4}{*}{Cora Full} 
        & G2P2 & \ding{55}  & 35.40{\scriptsize$\pm$5.63} & 27.45{\scriptsize$\pm$3.50} & 20.11{\scriptsize$\pm$7.29} \\
        & OFA & \ding{55} & 23.25{\scriptsize$\pm$5.15} & 33.40{\scriptsize$\pm$3.87} & 19.75{\scriptsize$\pm$3.73} \\
        \noalign{\vskip -2pt}  
        \cmidrule{2-6}
        \noalign{\vskip -2pt}  
        & STAG & \ding{51} & 47.50{\scriptsize$\pm$6.61} & 34.85{\scriptsize$\pm$4.89} & 61.58{\scriptsize$\pm$10.35} \\
        & STAG + $\mathbf{C}$ & \ding{55}  & \bf 62.80{\scriptsize$\pm$7.13} & \bf 61.80{\scriptsize$\pm$3.87} & \bf 73.17{\scriptsize$\pm$7.89} \\
        \bottomrule[1.2pt]
    \end{tabular}}
    \end{threeparttable}
    \begin{tablenotes}
    \footnotesize
    \item \textbf{Raw Feat +  Q}: Quantize raw node features into tokens for LLM inference.
    \item + $\mathcal{C}$: Classification using class-specific codebook.
    \end{tablenotes}
    \label{tab:zero_shot_rest}
\end{table}

\subsection{Zero-shot Performance}
\label{sec:appendix_zeroshot}
Table~\ref{tab:zero_shot_rest} presents zero-shot classification results on additional datasets (CiteSeer, PubMed, and ogbn-products). STAG with class-specific codebook maintains strong performance compared to raw feature baselines, demonstrating effective transfer of learned representations even without any target domain examples.

\begin{table}[t]
    \centering
    \footnotesize
    \caption{5-way 5-shot subgraph classification results (pre-trained on Cora Full).}
    \begin{threeparttable}
    \resizebox{0.9\linewidth}{!}{
    \begin{tabular}{l|c|c|c}
        \toprule[1.2pt]
        Method & Cora & Cora Full & Arxiv \\
        \midrule \midrule
        Raw Feat + Quantization & 67.75 & 78.32 & 65.18 \\
        STAG & \bf 69.60 & \bf 79.25 & \bf 68.41 \\
        \bottomrule[1.2pt]
    \end{tabular}}
    \end{threeparttable}
    \label{tab:subgraph_classification}
\end{table}

\subsection{Subgraph Classification}
\label{sec:appendix_task_generalization}
We evaluate 5-way 5-shot subgraph classification following L2P-GNN~\cite{lu2021learning}. We sample subgraphs around center nodes and use their labels for classification. Subgraph embeddings are obtained by mean-pooling node embeddings across all nodes within each subgraph, which are then quantized and used in prompts similar to our node classification approach.

As shown in Table~\ref{tab:subgraph_classification}, STAG consistently outperforms the raw feature quantization baseline across all three datasets. While the improvements are modest, they demonstrate that our structure-aware quantization approach successfully incorporates neighborhood information even at the subgraph level.

These comprehensive evaluations across link prediction, edge classification, and subgraph classification validate STAG's flexibility in handling different granularities and types of graph learning tasks beyond the primary focus on node classification.

\section{Few-shot and Zero-shot Prompt Templates}
\label{sec:appendix_prompt}
For few-shot classification, we use the following prompt template:

\begin{tcolorbox}[boxsep=0mm,left=2.5mm,right=2.5mm]
\small
You are an AI assistant tasked with classifying input word sequences into one of the following categories: [candidate classes are inserted here].

You must choose strictly from these categories and no others.

Each category has characteristic patterns shown in its examples.

Here are examples of input sequences and their corresponding categories to guide you:

[Support examples are inserted here]

When given a new input sequence, identify its key patterns and match them to the most similar category from the examples.

If no category is a clear match, choose the closest one.

**IMPORTANT:** Output only the category name and nothing else.

Input: [tokens of test node inserted here]
\end{tcolorbox}

For zero-shot classification, we use a simpler prompt template:

\begin{tcolorbox}[boxsep=0mm,left=2.5mm,right=2.5mm]
\small
You are an AI assistant tasked with classifying input word sequences into one of the following categories: [candidate classes are inserted here].

You must choose strictly from these categories and no others.

When given a new input sequence, classify it into one of the categories.

**IMPORTANT:** Output only the category name and nothing else.

Input: [tokens of test node inserted here]
\end{tcolorbox}

\end{document}